%% file: main.tex
\documentclass[sigconf,nonacm]{acmart}
\AtBeginDocument{%
  }

\setcopyright{acmlicensed}
\copyrightyear{2018}
\acmYear{2018}
\acmDOI{XXXXXXX.XXXXXXX}
\acmConference[Conference acronym 'XX]{Make sure to enter the correct
  conference title from your rights confirmation email}{June 03--05,
  2018}{Woodstock, NY}

\acmISBN{978-1-4503-XXXX-X/2018/06}

\usepackage{multirow}
\usepackage{algorithm}
\usepackage{algorithmic}
\usepackage[most]{tcolorbox}
\usepackage{pifont}
\usepackage{tikz}
\usepackage{makecell}
\usepackage{xcolor}
\usepackage{enumitem}
\usepackage[table]{xcolor}
\definecolor{oursblue}{HTML}{E6F3FF}

\definecolor{oursgray}{HTML}{E6E6E6}
\newcommand{\gcell}[1]{\cellcolor{oursgray}#1}
\definecolor{advgray}{RGB}{235,235,235}
\definecolor{raeblue}{RGB}{221,235,247}
\usepackage[capitalize,noabbrev]{cleveref}
\usepackage{subcaption}
\crefname{section}{Sec.}{Secs.}
\crefname{subsection}{Sec.}{Secs.} % optional: keep consistent for main text
\crefname{equation}{Eq.}{Eqs.}
\crefname{table}{Tab.}{Tabs.}
\crefname{figure}{Fig.}{Figs.}
\crefname{appendix}{App.}{Apps.}

\newcommand{\circlenode}[1]{%
    \resizebox{!}{0.8em}{%
        \tikz[baseline=(char.base)]{
            \node[shape=circle, fill=black, inner sep=0.8pt, text=white] (char) {#1};
        }%
    }%
}

\begin{document}

\title{Imperceptible and Reversible Adversarial Examples against Vision-Language Models for Privacy Protection}

\author{Qi Lu}
\email{luqi@hust.edu.cn}
\affiliation{%
  \institution{School of Cyber Science and Engineering, Huazhong University of Science and Technology} 
%   \postcode{430074}
%   \city{Wuhan}
%   \country{China}
  \country{}
}
% \authornote{Hubei Engineering Research Center on Big Data Security, School of Cyber Science and Engineering, HUST, Wuhan, 430074, China}
% \authornote{
%   National Engineering Research Center for Big Data Technology and System, Services Computing Technology and System Lab,  HUST, Wuhan, 430074, China
% \authornotemark[2]
% \authornotemark[3]
% \authornotemark[4]
% \authornote{National Engineering Research Center for Big Data Technology and System}
% \authornote{Services Computing Technology and System Lab}
% \authornote{Hubei Key Laboratory of Distributed System Security}
% \authornote{Hubei Engineering Research Center on Big Data Security}

\author{Ziqi Zhou}
\email{zhouziqi@cqu.edu.cn}
\affiliation{%
  \institution{College of Computer Science, Chongqing University} 
%   \postcode{430074}
%   \city{Wuhan}
%   \country{China}
  \country{}
}

\author{Yufei Song}
\email{yufei17@hust.edu.cn}
\affiliation{%
  \institution{School of Cyber Science and Engineering, Huazhong University of Science and Technology} 
%   \postcode{430074}
%   \city{Wuhan}
%   \country{China}
  \country{}
}
\authornote{Yufei Song is the corresponding author.}

\author{Zijing Li}
\email{lizijing@hust.edu.cn}
\affiliation{%
  \institution{School of Software and engineering, Huazhong University of Science and Technology} 
%   \postcode{430074}
%   \city{Wuhan}
%   \country{China}
  \country{}
}

\author{Lulu Xue}
\email{lluxue@hust.edu.cn}
\affiliation{%
  \institution{School of Cyber Science and Engineering, Huazhong University of Science and Technology} 
%   \postcode{430074}
%   \city{Wuhan}
%   \country{China}
  \country{}
}

\author{Minghui Li}
\email{minghuili@hust.edu.cn}
\affiliation{%
  \institution{School of Software Engineering, Huazhong University of Science and Technology} 
%   \postcode{430074}
%   \city{Wuhan}
%   \country{China}
\country{}
}

\author{Shengshan Hu}
\email{hushengshan@hust.edu.cn}
\affiliation{%
  \institution{School of Cyber Science and Engineering, Huazhong University of Science and Technology} 
%  \postcode{430074}
%  \city{Wuhan}
%  \country{China}
\country{}
}

\author{Leo Yu Zhang}
\email{leo.zhang@griffith.edu.au}
\affiliation{%
\institution{School of Information and Communication Technology, Griffith University}
%   \postcode{430074}
%   \city{Wuhan}
%   \country{China}
\country{}
}

\renewcommand{\shortauthors}{Qi Lu et al.}

\begin{abstract}
Vision–Language Models (VLMs) offer powerful multi-modal ability but also expose users to text-based privacy attacks where adversaries crawl online photos and query VLMs to extract sensitive attributes. Existing reversible adversarial example (RAE) methods protect images in purely visual tasks but fail in multi-modal settings, and current adversarial examples on VLMs rely on high-frequency noise that severely degrades visual quality. We propose CloakDiff, the first framework for reversible, high-fidelity privacy protection against text-based query attacks in VLMs. CloakDiff produces imperceptible adversarial examples by combining diffusion-based adversarial editing with an invertible network that embeds the original image for lossless recovery. It perturbs both pixel-space embeddings and manipulates latent cross attention maps to ensure strong cross-model and cross-prompt transferability while preserving global visual structure. To further enhance fidelity, we design EDM-Heuristic Sampling, a principled diffusion schedule for adversarial guidance. Experiments on multiple datasets and VLMs demonstrate that CloakDiff delivers multi-modal privacy preservation with high visual quality and reversibility. 
\end{abstract}




\setcopyright{none}
\settopmatter{printacmref=false}
\maketitle 
\input{sec/1_intro}

\input{sec/2_related}
\input{sec/3_preliminaries}
\input{sec/4-methodology}
\input{sec/5-experiment}
\input{sec/6-conclusion}
\bibliographystyle{ACM-Reference-Format}
\bibliography{sample-base}
\end{document}

%% file: sec/1_intro.tex
\section{Introduction}
\label{sec:intro}

\begin{figure}[!t]
    \centering
    \includegraphics[scale=0.46]{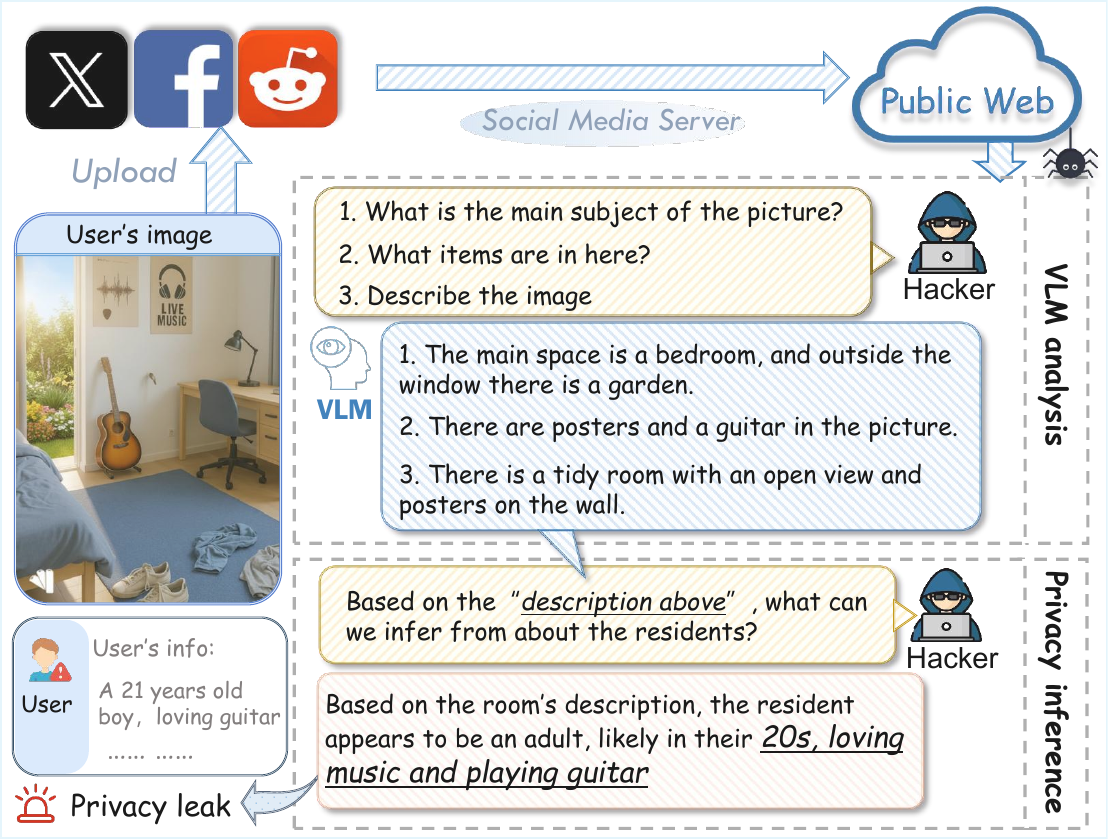}
    \caption{Illustration of privacy risks under VLMs.
    }
    \label{fig:intro}
    \vspace{-0.6cm}
\end{figure}

\textit{Vision-Language Models} (VLMs) ~\cite{li2023blip, liu2023visual,yu2025spa}  exhibit strong visual feature extraction and understanding capabilities, and are widely used in downstream tasks such as caption generation~\cite{fei2023transferable} and visual question answering (VQA)~\cite{khare2021mmbert}. 
However, when misused by malicious actors, they raise serious concerns about image privacy. 
Existing studies~\cite{vip} show that attackers can crawl user images and exploit VLMs to launch large-scale automated text queries that extract sensitive information.
Even benign-looking prompts can cumulatively expose private attributes such as gender, age, or location~\cite{staab2023beyond}.
As shown in \cref{fig:intro}, an attacker exploits a photo the user shared for decoration advice, obtains cues like “a poster” and “a guitar,” and infers the user is in his twenties~\cite{wei2022chain}.

To defend against such attacks, a practical privacy protection method should preserve the image’s utility for legitimate users while preventing unauthorized models from inferring private cues. Since the threat arises from machine perception rather than human perception, adversarial examples~\cite{zhou2023downstream,zhou2024securely,wang2025breaking} are a natural starting point: they can disrupt model predictions while introducing limited perceptual changes to the image~\cite{song2025seg,song2026erosion,yan2026vfacamou}. However, merely fooling the model is not sufficient for real-world privacy protection. In many scenarios, the original image still needs to be faithfully restored in trusted environments, making reversibility highly desirable. This motivates \textit{reversible adversarial examples} (RAEs)~\cite{zhu2024dp}, which combine adversarial protection with reversible data hiding to enable both privacy defense and lossless recovery.

\begin{figure}[!t]
    \centering
    \includegraphics[scale=0.50]{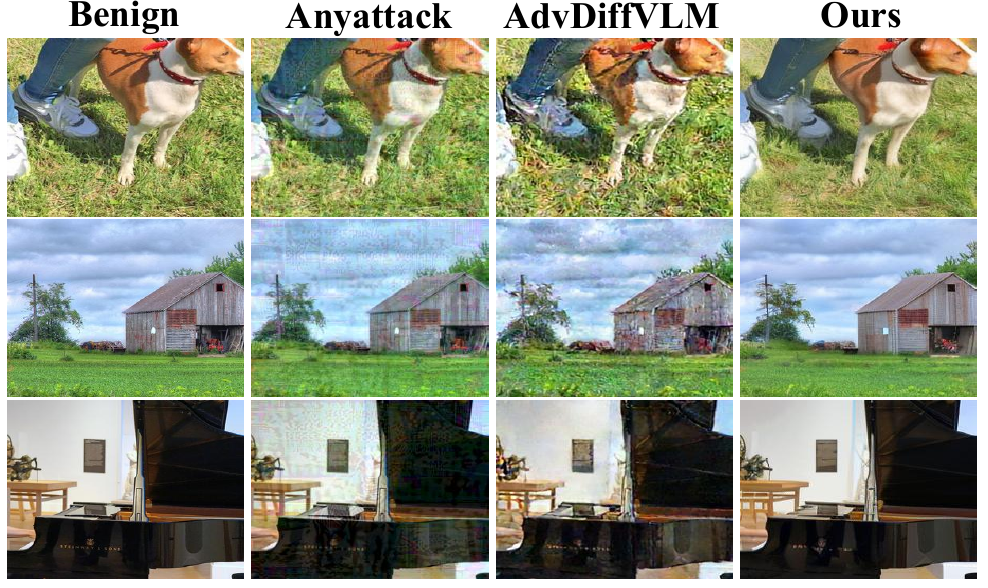}
    \caption{Comparison of adversarial attacks for VLMs.
    }
    \label{fig:visualize}
    \vspace{-0.6cm}
\end{figure}

A natural question therefore is whether adversarial examples can be repurposed as a practical privacy protection mechanism against VLMs. Prior studies suggest a promising starting point: VLMs are vulnerable to adversarial examples. AttackVLM~\cite{zhao2023evaluating} introduced transferable black-box attacks on VLMs, and Anyattack~\cite{zhang2024anyattack} further improved attack performance with a self-supervised perturbation generator. These advances demonstrate that adversarial examples can effectively mislead VLMs. However, they remain inadequate for privacy protection in practice, as they typically introduce high-frequency~\cite{zhou2025numbod,zhou2024darksam} perturbations that noticeably degrade image quality (see \cref{fig:visualize}). In addition, privacy protection in multimodal settings must remain effective under diverse prompts and across different model architectures. Therefore, the central challenge is:

\begin{quote}
    \emph{How to achieve cross-model and cross-prompt privacy preservation while maintaining high visual quality?}  
\end{quote}

Inspired by prior studies on \textit{unrestricted adversarial examples} (UAEs)~\cite{jia2022adv, yuan2022natural}, which are capable of producing adversarial examples without noticeable noise while maintaining high visual quality, we explore their potential for privacy protection in VLMs.
AdvDiffVLM~\cite{guo2024efficient} first applied UAE to VLMs, using a set of visual encoders from CLIP ~\cite{zhou2023advclip} as surrogate models and ensemble gradient estimation to enhance transferability. 
However, the heavy disruption to the sampling process produces low-quality images, as shown in \cref{fig:visualize}. Moreover, its end-to-end generation lacks explicit noise encoding, rendering existing RAE methods inapplicable.

In this work, we propose CloakDiff, the first privacy-preserving framework designed to defend against text-based query attacks in VLMs. Given a user-uploaded image, CloakDiff sequentially processes it through a diffusion model~\cite{wangadvedm} and an Invertible Neural Network (INN) ~\cite{ardizzone2018analyzing, jing2021hinet}, producing imperceptible yet reversible adversarial examples. CloakDiff exhibits three key properties:
\circlenode{1} \textbf{Cross-model and cross-prompt privacy protection.} CloakDiff injects adversarial semantics into the sampling process via adversarial guided editing. In the pixel domain, CloakDiff shifts the original embeddings to disrupt  semantic representations, while in the latent domain it alters cross attention maps to achieve fine-grained disruption of multi-modal alignments, thereby enhancing transferability.
\circlenode{2} \textbf{High visual fidelity.} CloakDiff preserves the global structure by maintaining the self-attention maps during sampling, ensuring visual fidelity and perceptual quality.
\circlenode{3}  \textbf{Reversibility.} CloakDiff treats the original image as secret image and adversarial example as cover image to generate RAE. In revealing phase, the RAE alone suffices to recover the original image.
Furthermore, we introduce \textit{EDM-Heuristic Sampling} (EHS), a tailored diffusion schedule for adversarial guided sampling. EHS improves image quality while providing a principled formulation from the perspective of score matching. Our main contributions are summarized as follows:

\begin{itemize}[leftmargin=1.2em]
\item 
We present the first defense against text-based query attacks in vision-language models by generating imperceptible and reversible adversarial examples for image privacy protection.

\item 
We leverage a diffusion model for adversarial example generation by manipulating attention maps during sampling, and further develop a customized scheduling strategy for adversarially guided editing with theoretical justification.

\item 
Experiments on multiple datasets and models show that CloakDiff achieves strong cross-model and cross-prompt effectiveness while preserving high visual fidelity and reversibility.
\end{itemize}

%% file: sec/2_related.tex
\section{Related Works}
\label{sec:related}

\subsection{Adversarial Examples against VLMs}
VLMs initially align visual and textual embeddings in a shared representation space. 
Subsequent models introduce fusion mechanisms such as query transformers~\cite{li2023blip} and cross attention modules~\cite{alayrac2022flamingo}, while recent systems integrate large-scale vision encoders~\cite{chen2024internvl} into pretrained LLMs~\cite{liu2023visual}. 
Robustness evaluation of VLMs increasingly centers on adversarial attacks~\cite{wang2025breaking,song2025segment,zhou2025sam2,yan2026transferable,zhu2026transferable}. Prior work~\cite{zhang2022towards, zhou2023advclip} studies adversarial robustness in classification and retrieval scenarios. AttackVLM~\cite{zhao2023evaluating} introduces the first query-based black-box attack for VQA, motivating later research on black-box transfer attacks. Among these, Anyattack~\cite{zhang2024anyattack} reports the strongest performance by pretraining a noise generator. AdvDiffVLM~\cite{guo2024efficient} generates UAEs using a diffusion model tailored for VLMs, but the perturbations substantially degrade visual fidelity. Consequently, existing attacks often produce visually distorted examples, limiting their practicality in privacy-preserving applications.

\subsection{Image Steganography}
Image steganography hides a secret image within a cover image while ensuring both imperceptibility and recoverability. 
Recent work adapts this idea to construct RAEs by embedding adversarial noise that can later be removed to restore the original image. 
Compared with traditional techniques, deep learning–based steganography offers higher capacity and stronger concealment~\cite{sanjalawe2025deep}. 
\textit{Invertible neural networks} (INNs)~\cite{radev2020bayesflow} are particularly well suited to image hiding because they jointly model forward and inverse mappings through shared parameters. Hi-Net~\cite{jing2021hinet} demonstrates the feasibility of hiding full-resolution images using INNs. However, to our knowledge, no prior work investigates leveraging deep steganography to build reversible adversarial examples.

%% file: sec/3_preliminaries.tex
\section{Problem Statement}
\label{sec:preliminary}

\vspace{0.1cm}
\begin{figure*}[!t]
    \centering
    \includegraphics[scale=0.60]{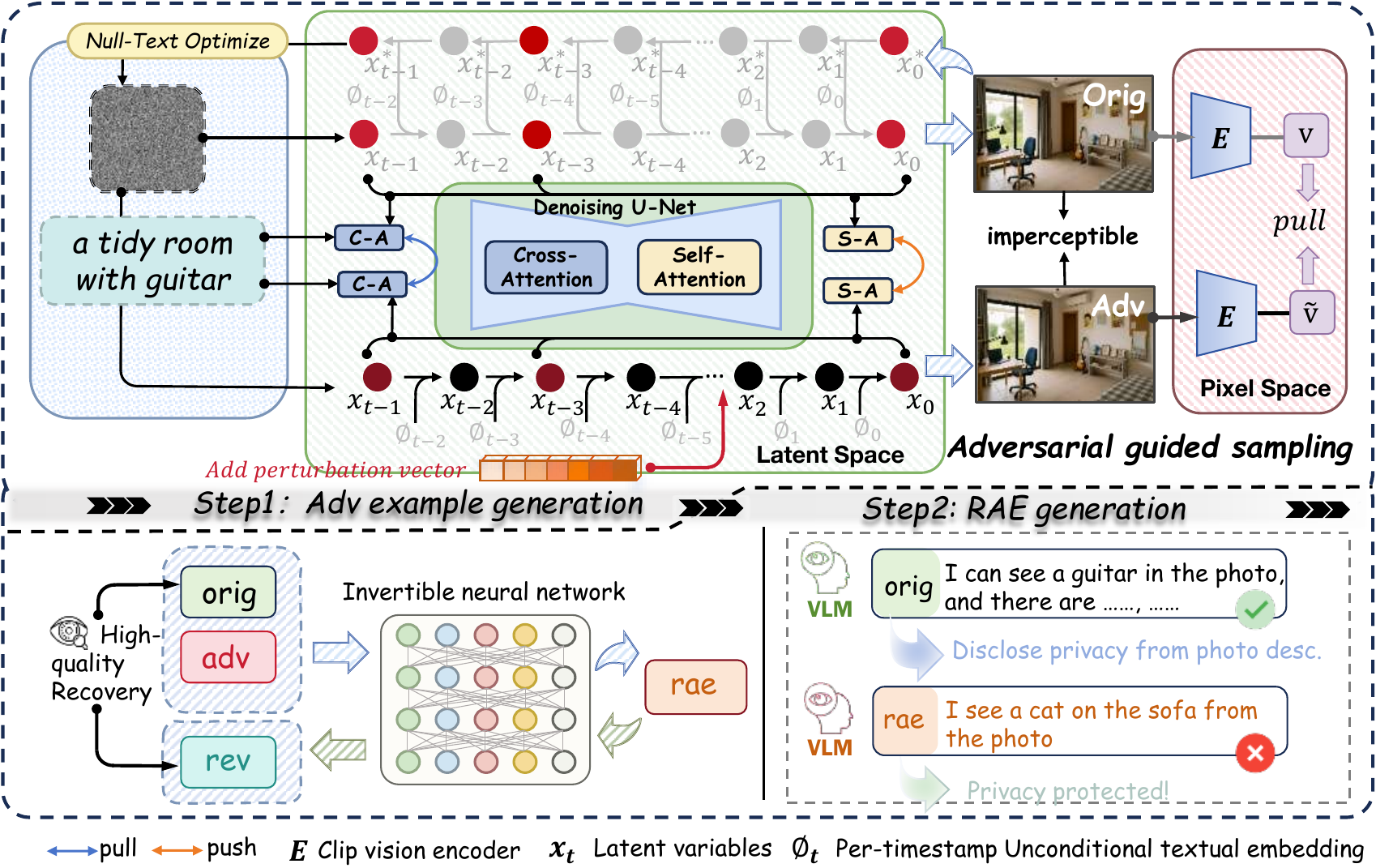}
    % \vspace{-15pt}
    \caption{The framework of CloakDiff: the first privacy-preserving method for image protection under VLMs.
    }
    \label{fig:pipeline}
    \vspace{-0.4cm}
\end{figure*}

We define image privacy as: Information in an image that can be either directly identified from visual content or indirectly inferred from visual cues, scene context, and external world knowledge, and ultimately linked to an identified or identifiable natural person~\cite{vip}. 

Unlike conventional visual privacy settings~\cite{li2024transferable} that focus on the explicit presence of people, this definition also encompasses inferential privacy leakage arising from non human subjects, surrounding objects, and traces of daily life. Under this definition, we consider a user-side privacy protection scheme with black-box access to the attacker’s VLM and prompts. Given an original image $x_{\mathrm{orig}}$, the scheme produces a privacy-protected image $x_{\mathrm{pro}}$, satisfying three below key properties:

\textbf{(I) $x_{\mathrm{pro}}$ should ensure cross-model and cross-prompt privacy protection.} It should consistently suppress private attribute inference under unseen attack models and arbitrary natural language prompts, so that an attacker cannot reliably recover sensitive user attributes even when varying the VLM architecture, alignment scheme, reasoning strategy, or query prompt.

\textbf{(II) $x_{\mathrm{pro}}$ should preserve high visual fidelity.} It should remain visually consistent with $x_{\mathrm{orig}}$ while retaining natural appearance, semantic integrity, and perceptual quality, thereby remaining suitable for routine scenarios such as browsing, sharing, uploading, and compressed transmission.

\textbf{(III) $x_{\mathrm{pro}}$ should support direct recovery of $x_{\mathrm{orig}}$.} The user should be able to reconstruct a high quality image $x_{\mathrm{orig}}^{\mathrm{rev}}$ from $x_{\mathrm{pro}}$ alone, without any auxiliary information, thereby enabling reversible privacy protection from public release back to trusted use.

%% file: sec/4-methodology.tex
\section{Methodology}
\label{sec:methodology}

In multi-modal scenarios, existing adversarial example generation methods typically inject high-frequency noise directly into images, which substantially degrades perceptual quality~\cite{zhang2024anyattack}. 
We therefore leverage diffusion models~\cite{rombach2022high} to generate imperceptible adversarial images. Specifically, we perform feature alignment in the pixel space with a pretrained encoder, and control cross attention in the latent space to enable \textit{adversarial guided editing}, thereby injecting adversarial semantics into the original image during sampling. Furthermore, to better preserve image structure and fidelity, we retain self attention map and introduce a tailored sampling schedule. The overall pipeline is illustrated in \cref{fig:pipeline}.

\subsection{Adversarial Guided Editing}
Diffusion models generate images by progressively denoising latent variables during the sampling process, while generating adversarial examples involves gradually injecting adversarial information into this process. Unlike conventional image editing methods ~\cite{kawar2023imagic, liu2024towards} that follow user textual instructions, our approach focuses on injecting adversarial information into the sampling process. We call this process as adversarial guided editing. Adversarial guided editing operates jointly in the pixel space and latent space, progressively injecting adversarial semantics via a score-matching strategy. Given an original image $x_{\mathrm{orig}}$, the edited output is denoted as $x_{\mathrm{adv}}$.

The pixel-space guidance dominates the adversarial editing process. Pretrained encoder provides a unified embedding space for image-text alignment, a natural way to degrade semantics is to shift the encoder feature embedding of $x_{\mathrm{adv}}$ away from that of $x_{\mathrm{orig}}$. Empirically, we find that aligning $x_{\mathrm{adv}}$ toward a target embedding is more effective than merely diverging from its original representation. Let $f_{\phi}(\cdot)$ denote the CLIP image encoder. The pixel-space adversarial guidance $\mathcal{L}_\mathrm{pixel}$ is formulated as:
\begin{equation}
\mathcal{L}_\mathrm{pixel}= -\;
\frac{f_{\phi}(x_{\mathrm{adv}})}{\|f_{\phi}(x_{\mathrm{adv}})\|_2}^\top \cdot \; \frac{f_{\phi}(x_{\mathrm{tgt}})}{\|f_{\phi}(x_{\mathrm{tgt}})\|_2}\;,
\label{eq:eq6}
\end{equation}
where $x_{\mathrm{tgt}}$ is the target image as a shift anchor (see ~\cref{sec:experiments} for details).

\begin{figure}[t]
    \centering
    \subcaptionbox{Activation of $x_\mathrm{orig}$}[0.23\textwidth]{%
        \includegraphics[width=0.23\textwidth]{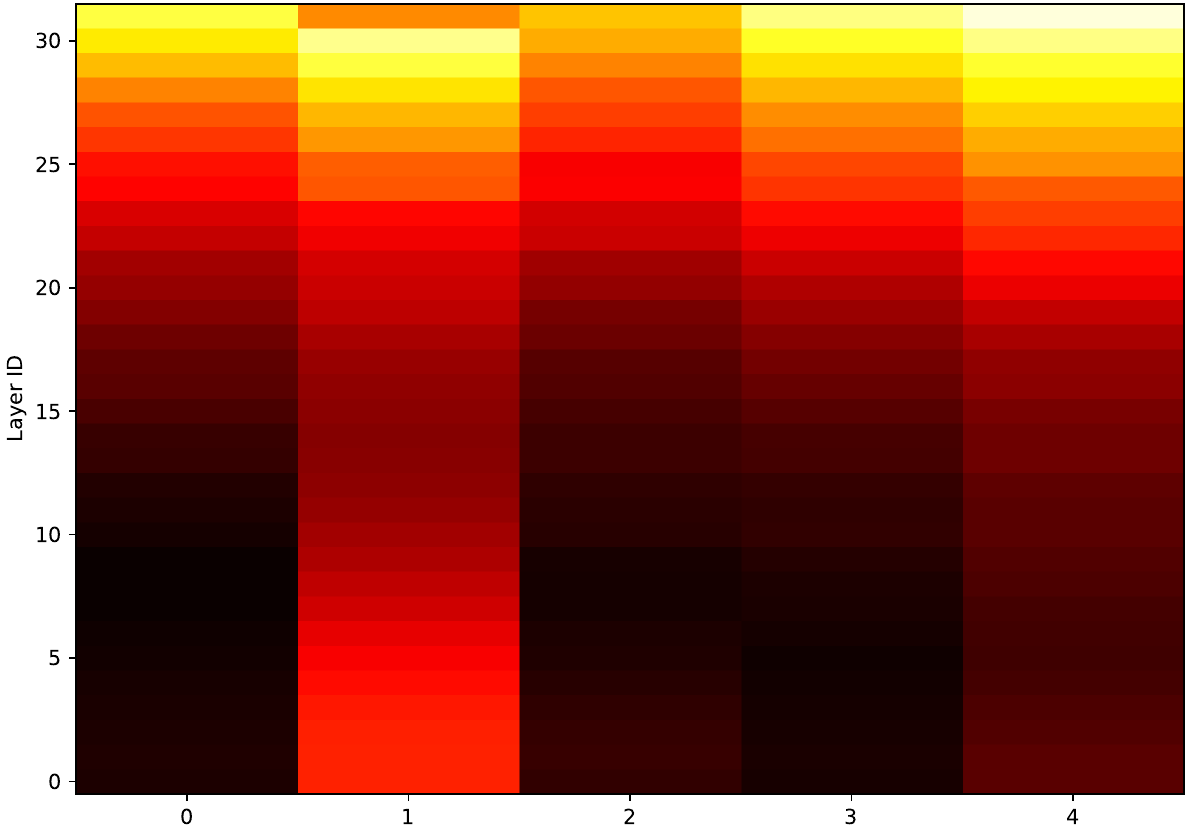}%
    }\hfill
    \subcaptionbox{Activation of $x_\mathrm{adv}$}[0.24\textwidth]{%
\includegraphics[width=0.23\textwidth,height=2.85cm,keepaspectratio=false]{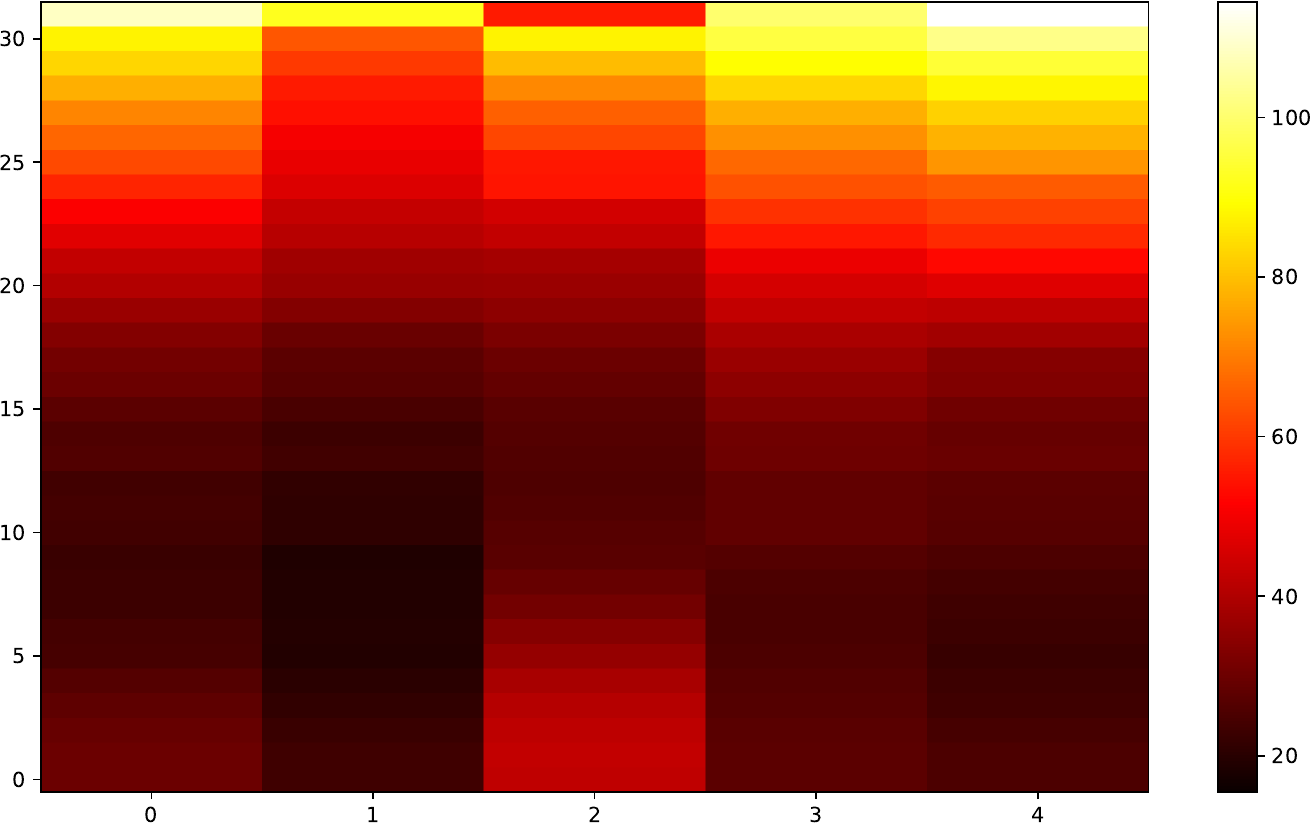}%
    }
    \caption{Visualization of hidden states in LLaVA \cite{liu2023visual}, revealing how visual and textual modalities interact. The y-axis is the final-token activation across layers during inference, and the colors indicate activation values.}
    \label{fig:cross}
    \vspace{-4mm}
\end{figure}

While pixel-space guidance directly disrupts semantics, latent-space guidance perturbs the underlying multi-modal alignment more delicately. Previous work ~\cite{liu2024towards} shows that manipulating cross attention maps in the latent space effectively enables textual instructions to control image semantics. 
Inspired by this, we perturb the cross attention maps during sampling to disrupt the fine-grained alignment between image and text modalities. As shown in \cref{fig:cross}, this strategy significantly impairs multi-modal consistency even under black-box VLMs, demonstrating strong transferability. 
Diffusion model adopts \textit{Classifier-Free Guidance} (CFG) ~\cite{ho2022classifier} to modify the noise prediction by interpolating between unconditional and conditional outputs:
\begin{equation}
\epsilon_\theta^{\rm cfg}(x_t) = \epsilon_\theta(x_t\mid \varnothing) + g\Bigl(\epsilon_\theta(x_t\mid c)-\epsilon_\theta(x_t\mid \varnothing)\Bigr)\;,
\label{eq:cfg}
\end{equation}
where $g$ is the guidance scale. Following \cref{eq:cfg}, the conditional branch is guided by $c$, a text prompt generated via a pretrained captioner ~\cite{mokady2021clipcap}. 
To preserve essential caption semantics, we employ the template strategy proposed in ~\cite{xie2025chain}. Based on the conditional noise prediction, we define the latent-space adversarial guidance :
\begin{equation}
\mathcal{L}_\mathrm{latent} = -\;\Big\|
\frac{1}{L^{c} \cdot H^{c}}\sum_{l=1}^{L^{c}}\sum_{h=1}^{H^{c}} \left(A^{l,h}_{\mathrm{c}}(x_{\mathrm{adv}}) - A^{l,h}_{\mathrm{c}}(x_{\mathrm{orig}})\right)\Big\|_2\;,
\label{eq:eq7}
\end{equation}
where $L^{c}$ and $H^{c}$ respectively denote the number of cross attention layers and heads, and $A^{l,h}_{\mathrm{cross}}(x)$ represents the cross attention map at layer $l$, head $h$ during the generation. \cref{eq:eq7} enforces a precise deviation between the cross attention maps of $x_{\mathrm{adv}}$ and those of $x_{\mathrm{orig}}$, thereby corrupting multi-modal semantics. To enhance reconstruction precision, we adopt \textit{Null-Text Optimization} (NTO):
\begin{equation}
\hat{\varnothing} = \varnothing - \alpha_1
\frac{\partial}{\partial \varnothing}
\big|x_{\mathrm{rever}} - x_{\mathrm{recon}}(\varnothing)\big|_2^2\;,
\label{eq:eq8}
\end{equation}
where $\alpha_1$ is the update step size, $\hat{\varnothing}$ denotes the optimized unconditional text embedding, and $x_{\mathrm{rever}}$, $x_{\mathrm{recon}}(\varnothing)$ represent the original and reconstructed latent variables.

\subsection{Self-Attention Based Retention}
As discussed in \cref{sec:intro}, a major challenge in generating $x_{\mathrm{adv}}$ with diffusion models lies in maintaining visual fidelity while injecting adversarial semantics. Prior work~\cite{wang2021feature} shows that the transferability of adversarial examples mainly stems from perturbing key regions, rather than making brute-force changes across the entire image. Motivated by this, we introduce a self-attention based visual retention mechanism that constrains adversarial guided editing to local key regions while preserving global structure. This design improves perceptual quality without sacrificing protection performance.

Self-attention mechanism plays a central role in coupling edited and non-edited regions ~\cite{liu2024towards} in diffusion model. By regulating self-attention maps, we can achieve fine-grained local editing while maintaining global consistency. We therefore add a visual retention constraint as a compensatory term $\mathcal{L}_\mathrm{retain}$ to adversarial guided editing. This term encourages the self-attention maps during sampling to remain close to those of the original reconstruction, balancing visual fidelity and adversarial effectiveness:
\begin{equation}
\mathcal{L}_\mathrm{retain} = \;\Big\|
\frac{1}{L^{s} \cdot H^{s}}\sum_{l=1}^{L^{s}}\sum_{h=1}^{H^{s}} \left(A^{l,h}_{\mathrm{s}}(x_{\mathrm{adv}}) - A^{l,h}_{\mathrm{s}}(x_{\mathrm{orig}})\right)\Big\|_2\;.
\label{eq:eq9}
\end{equation}
Here, $A^{l,h}_{\mathrm{self}}(\cdot)$ represents the self-attention map at layer $l$, head $h$. 

% By preserving the structural similarity of self-attention maps, this constraint effectively retains the global semantics of $x_{\mathrm{orig}}$ while allowing localized adversarial manipulation, leading to higher visual realism and more stable protection.

\subsection{EDM-Heuristic Sampling}

Sampling schedules~~\cite{salimans2022progressive} play a critical role in the quality of images generated by diffusion models. 
Diffusion models generate data by gradually denoising a noisy sample. At each step $t$, a scaling factor $\alpha_t \in (0,1)$ controls the noise level. From the score-matching perspective~\cite{song2020score}, the model learns the gradient of the log-density of noisy data, i.e., the score function $s_\theta(x_t,t)$. The reverse sampling step is then formulated as:
\begin{equation}
x_{t-1} = \frac{1}{\sqrt{\alpha_t}}\left(x_t + (1-\alpha_t)s_\theta(x_t,t)\right) + \sigma_t z\;,
\label{eq:sample}
\end{equation}
where $z \sim \mathcal{N}(0,I)$ introduces stochasticity, and $\sigma_t$ controls the noise level. In our context, adversarial guided editing modifies $x_t$, thereby affecting both the trajectory of the sampling process and the resulting image quality, highlighting the need for customized schedules tailored to adversarial guided sampling.

\begin{figure}[!t]
    \centering
    \includegraphics[scale=0.20]{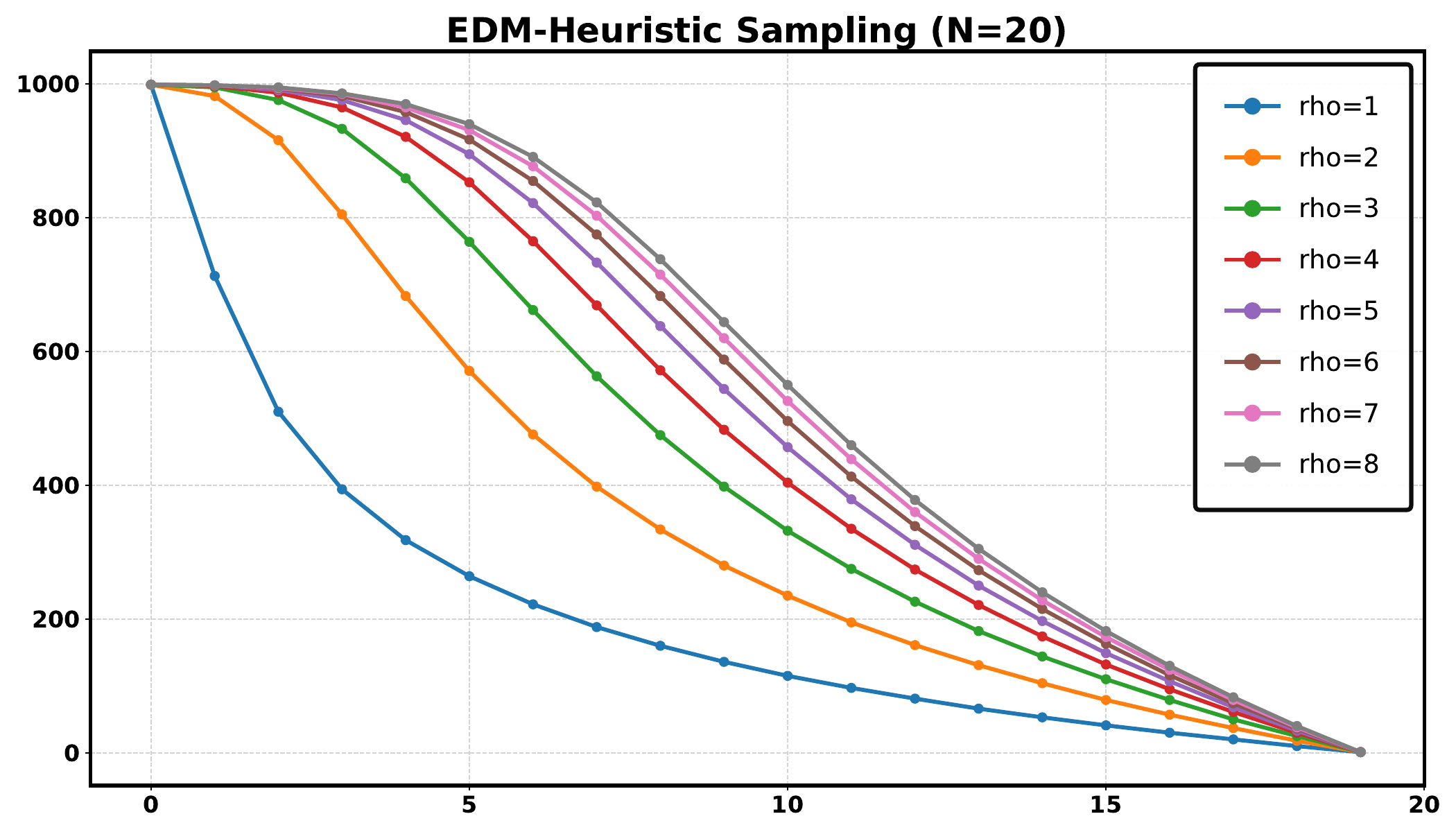}
    \caption{This figure illustrates the positions in the model’s original latent space corresponding to different EHS sampling steps. Sampling schedules under different $\rho$ control the density of $\bar{\alpha}_t^{\mathrm{syn}}$ across different noise levels: a larger $\rho$ yields smaller selected values of $\bar{\alpha}_t^{\mathrm{syn}}$, consistent with Theorem I.
    }
    \label{fig:edm}
    \vspace{-0.4cm}
\end{figure}

To address this issue, we draw inspiration from EDM~\cite{karras2022elucidating}, which decouples diffusion training from sampling and allows training-free schedule design. Building on this insight, we propose \textit{EDM-Heuristic Sampling} (EHS), an adaptive and nonlinear time discretization strategy. Intuitively, under the same total number of sampling steps, EHS tends to select larger values of $\bar{\alpha}_t$ (~\cref{eq:sample}), corresponding to latent states with lower noise. Since the derivative of the reverse \textit{ordinary differential equation} (ODE) varies rapidly for large values of $\bar{\alpha}_t$ in the $\alpha$-space ~\cite{karras2022elucidating}, leading to exponential amplification of sampling errors. Denser sampling is required to accurately inject adversarial information under such high-error conditions, consistent with the design philosophy of EDM. We transform the noise scale as $u=\sigma^{1/\rho}$, sample uniformly in $u$, and map the samples to the \textit{signal-to-noise} (SNR) domain:
\begin{equation}
\bar{\alpha}_t = \frac{1}{1 + \left(u_{\max} + \frac{t}{N-1}(u_{\min} - u_{\max})\right)^{2\rho}}\;,
\label{eq:eq10}
\end{equation}
where $u_{\max} = \sigma_{\max}^{1/\rho}$, $u_{\min} = \sigma_{\min}^{1/\rho}$. $N$ is the total number of sampling steps and $\,t\,$ range from $\,0\,$ to $\,N -1\,$, with $\,\rho\,$ controls the convexity of the schedule.
To adjust the convexity of the sequence, we apply a symmetric inversion with linear interpolation to the $\bar{\alpha}_t$ sequence to obtain the final series $\bar{\alpha}_t^{\mathrm{syn}}$. This yields a nonlinear, density-controlled time discretization based on $\bar{\alpha}_t^{\mathrm{syn}}$, as shown in \cref{fig:edm}.

\textbf{Theorem I}: \,\,Let $\bar{\alpha}_t^{\mathrm{syn}}(\rho)$ denote the EHS sequence as a function of the exponent parameter $\rho$. Then, for each fixed $\,t$, $\bar{\alpha}_t^{\mathrm{syn}}(\rho)$ is monotonically decreasing in $\rho$:
% \vspace{-0.2cm}
$$
\text{if } \rho_1 < \rho_2, \quad \bar{\alpha}_t^{\mathrm{syn}}(\rho_1) \ge \bar{\alpha}_t^{\mathrm{syn}}(\rho_2).
$$

This theorem enables users to control the noise level at each editing step via a single parameter $\rho$, allowing for flexible adaptation. From the perspective of score matching, EHS describes adversarial guided sampling as:
\begin{equation}
s_\theta^{\mathrm{\mathrm{adv}}}(\tilde{x}_t, t) = s_\theta (\tilde{x}_t, t) -  \alpha_2 \sum_{k} \lambda_k \nabla_{\tilde{x}_t} \mathcal{L}_k(\tilde{x}_t)\;,
\label{eq:eq11}
\end{equation}
% \vspace{-0.4cm}
\begin{equation}
\tilde{x}_{t-1} = \frac{1}{\sqrt{\alpha_t}}\Big(\tilde{x}_t + (1-\alpha_t)s_\theta^\mathrm{\mathrm{adv}}(\tilde{x}_t, t)\Big) + \sigma_t z\;,
\label{eq:eq12}
\end{equation} where $k \in \{\mathrm{pixel},\ \mathrm{latent},\ \mathrm{attn}\}$, $\lambda_k$ are the corresponding weighting coefficients, $\alpha_2$ scales the adversarial term, and $\tilde{x}_t$ denotes the perturbed latent variable.

\definecolor{oursblue}{HTML}{E6F3FF} 
\begin{table*}[t!]
    \caption{CloakDiff’s protection performance evaluated on nine VLMs and two datasets. For all metrics in the table, smaller values indicate stronger protection performance. To unify the scale, all relevant metrics are normalized to the 0–100.}
    \vspace{-0.2cm}
  \centering

% \scalebox{0.8}{
    \begin{tabular}{ccccccccccccc}
    \toprule[1.5pt]
    \multirow{2}[0]{*}{\textbf{Model}}  & \multicolumn{6}{c}{\textbf{ImageNet}}                  & \multicolumn{6}{c}{\textbf{MS-COCO}} \\
\cmidrule(lr){2-7} \cmidrule(lr){8-13}      & BLEU-1 & BLEU-4 & CIDEr & METEOR & ROUGEL & SPICE & BLEU-1 & BLEU-4 & CIDEr & METEOR & ROUGEL & SPICE \\ \cmidrule(lr){1-13} 
    BLIP  & 64.25 & 44.10  & 41.91 & 59.17 & 63.23 & 45.18 & 77.92 & 60.24 & 56.54 & 73.71 & 75.41 & 61.91 \\
    BLIP-2 & 47.71 & 35.59 & 27.21 & 47.31 & 48.04 & 44.75 & 46.50  & 34.49 & 30.33 & 46.85 & 49.73 & 44.54 \\
    InstructBLIP & 56.29 & 33.36 & 23.31 & 47.26 & 48.51 & 36.38 & 57.89 & 36.22 & 25.07 & 49.06 & 50.35 & 42.09 \\
    Flamingo & 34.27 & 16.15 & 6.91 & 28.74 & 28.98 & 23.10 & 39.74 & 18.88  & 6.34  & 32.13 & 32.49 & 28.66 \\
    Unidiffuser & 37.10  & 16.29 & 13.57 & 31.85 & 34.88 & 14.86 & 36.35 & 13.09 & 11.40  & 28.87 & 33.16 & 16.32 \\
    LLaVA & 53.09 & 31.05 & 21.51 & 44.48 & 45.61 & 32.29 & 59.41 & 37.47 & 26.01 & 50.95 & 51.85 & 41.92 \\
    MiniGPT-4 & 39.41 & 10.74 & 17.96 & 27.82 & 26.34 & 19.96 & 39.88 & 11.60  & 22.23 & 25.83 & 25.74 & 20.18 \\
    Qwen  & 51.03 & 27.09 & 21.44 & 47.84 & 46.48 & 42.98 & 48.19 & 23.00    & 16.34 & 43.03 & 41.10  & 38.24 \\
    Internvl & 48.06 & 23.91 & 16.51 & 40.21 & 40.93 & 35.94 & 46.58 & 21.43 & 14.50  & 40.70  & 40.11 & 29.42 \\
    \bottomrule[1.5pt]
    \end{tabular}
% }
  \label{tab:main}
  \vspace{-0.4cm}
\end{table*}

\subsection{INN-Based Reversible Steganography}
% Through adversarial guidance sampling, the user-uploaded image $x_{\mathrm{orig}}$ is transformed by the diffusion model into an imperceptible adversarial example $x_{\mathrm{adv}}$. 
Existing RAE methods ensure reversibility by embedding explicit noise representations into the cover image. In contrast, $x_{\mathrm{adv}}$ is produced through a generative diffusion process rather than direct perturbation of $x_{\mathrm{orig}}$, and thus lacks an explicit noise encoding. Consequently, conventional methods cannot be directly applied for reconstruction. To address this limitation, we introduce an INN-based reversible steganography method that embeds $x_{\mathrm{orig}}$ directly into $x_{\mathrm{adv}}$ using an invertible neural network. This design enables high quality recovery of the original image while maintaining visual fidelity. Specifically, we adopt the architecture proposed in ~\cite{jing2021hinet}, denoted as $F_\mathrm{INN}$, which jointly models the concealing and revealing phases as the forward and inverse propagations of the same network. During the concealing stage, $F_\mathrm{INN}$ takes the adversarial image $x_{\mathrm{adv}}$ and the original image $x_{\mathrm{orig}}$ as inputs to produce a reversible adversarial example $x_{\mathrm{rae}}$, which ultimately serves as the privacy protected image $x_{\mathrm{pro}}$:
\begin{equation}
x_{\mathrm{pro}} = x_{\mathrm{rae}} = F_\mathrm{INN} (x_{\mathrm{adv}},\; x_{\mathrm{orig}})\;.
\label{eq:eq13}
\end{equation}
The generated $x_{\mathrm{rae}}$ is visually imperceptible from $x_{\mathrm{orig}}$ and achieves high perceptual quality. In the revealing phase, the original image can be exactly reconstructed from $x_{\mathrm{rae}}$ alone by applying the inverse propagation of $F_\mathrm{INN}$:
\begin{equation}
x^{\mathrm{rev}}_{\mathrm{orig}} = F_\mathrm{INN}^{-1}(x_{\mathrm{rae}})\;.
\label{eq:eq14}
\end{equation}

In practice, recovered $x^{\mathrm{rev}}_{\mathrm{orig}}$ closely approximates $x_{\mathrm{orig}}$ (see~\cref{tab:fidelity}), ensuring high quality reversibility.

\begin{table}[t!]
  \caption{Comparison of visual quality and recovery quality. ‘CloakDiff-ADV’ denotes $x_{\mathrm{adv}}$ without steganography, so recovery quality is unavailable (‘—’), while ‘CloakDiff-RAE’ denotes $x_{\mathrm{rae}}$. Blue indicates the best result, and gray indicates the second-best. The same with~\cref{tab:comparison_protection_2x2}.}
  \vspace{-0.2cm}
  \centering
  \renewcommand{\arraystretch}{0.9}
    \setlength{\tabcolsep}{8pt}
  \resizebox{\columnwidth}{!}{%
  \begin{tabular}{lccccc}
    \toprule[1.5pt]
    \multirow{2}[3]{*}{\textbf{Method}} & \multicolumn{2}{c}{\textbf{Visual Quality}} & \multicolumn{3}{c}{\textbf{Recovery Quality}} \\
    \cmidrule(lr){2-3} \cmidrule(lr){4-6}
    & \textbf{LPIPS} & \textbf{FID} & \textbf{SSIM} & \textbf{PSNR} & \textbf{RMSE} \\
    \cmidrule(lr){1-6}
    DP-RAE~\cite{zhu2024dp}              & 14.37 & 93.27  & 0.97 & 37.84 & 4.20 \\
    Anyattack~\cite{zhang2024anyattack} & 30.18 & 159.03 & 0.98 & 42.08 & 0.0014 \\
    AttackVLM~\cite{zhao2023evaluating} & 13.54 & 85.29  & 0.98 & 41.20 & 0.0021 \\
    AdvDiffVLM~\cite{guo2024efficient}  & 20.90 & 133.38 & 0.99 & 44.38 & 0.0004 \\
    \rowcolor{advgray}
    \gcell\textbf{CloakDiff-ADV}              & 10.76 & 73.08  & ——   & ——    & —— \\
    \rowcolor{raeblue}
    \textbf{CloakDiff-RAE}              & 10.85 & 73.09  & 0.99 & 44.37 & 0.0004 \\
    \bottomrule[1.5pt]
  \end{tabular}
  }
  \label{tab:fidelity}
  \vspace{-0.6cm}
\end{table}

\renewcommand{\algorithmicrequire}{\textbf{Input:}} 
\renewcommand{\algorithmicensure}{\textbf{Output:}} 
\input{sec/algo}

%% file: sec/algo.tex
\begin{algorithm}[H]
    \caption{\;CloakDiff}
    \label{algorithm}
    \begin{algorithmic}[1]
        \REQUIRE User's image $x_{\mathrm{orig}}$, target image $x_{\mathrm{tgt}}$, pretrained diffusion model, CLIP encoder $f_{\phi}$, INN model $F_\mathrm{INN}$, total sampling steps $N$, EHS parameter $\rho$, loss weights $\{\lambda_k\}_{k \in \{\mathrm{pixel}, \mathrm{latent}, \mathrm{retain}\}}$, step sizes $\alpha_1$, adversarial scale $\alpha_2$, conditioned caption $c$, total iteration epochs $K$
        \ENSURE $x_{\mathrm{rae}}$
        
        \STATE \textbf{// Phase 1: Adversarial Guided Sampling}
        \STATE Compute $\{\bar{\alpha}_t^{\mathrm{syn}}\}_{t=0}^{N-1}$ via Eq.~\eqref{eq:eq10} with parameter $\rho$
        % \STATE Generate caption $c$ for $x_{\mathrm{orig}}$ using pretrained captioner
        % \STATE Apply template strategy to preserve essential semantics: $c \leftarrow \text{Template}(c)$
        \STATE Encode $x_{\mathrm{orig}}$ to latent and Initialize $\tilde{x}_{N-1}$
        % \STATE Sample noise: $z_T \sim \mathcal{N}(0, I)$
        % \STATE \textbf{// Null-Text Optimization for accurate reconstruction}
        % \STATE Initialize unconditional embedding: $\varnothing \leftarrow \text{NullText}$
        \STATE Optimize unconditional embedding ${\varnothing}$ via Eq.~\eqref{eq:eq8}
        % to minimize $|x_{\mathrm{rever}} - x_{\mathrm{recon}}(\varnothing)|_2^2$
        
        % \STATE \textbf{// EHS: Generate adaptive sampling schedule}
        
        % \STATE \textbf{// Adversarial guided sampling}        
        % \leftarrow z_T$
        \FOR{$\,i = K$ \textbf{to} $\,1\,$}
        \FOR{$\,t = N - 1$ \textbf{to} $\,0\,$}
            % \STATE Compute conditional noise: $\epsilon_{\mathrm{c}} \leftarrow \epsilon_\theta(\tilde{x}_t, t, c)$
            % \STATE Compute unconditional noise: $\epsilon_{\mathrm{u}} \leftarrow \epsilon_\theta(\tilde{x}_t, t, \hat{\varnothing})$
            % \STATE Apply classifier-free guidance: $\epsilon_{\mathrm{cfg}} \leftarrow \epsilon_{\mathrm{u}} + w \cdot (\epsilon_{\mathrm{c}} - \epsilon_{\mathrm{u}})$
            
            % \STATE \textbf{// Extract attention maps during denoising}
            \STATE Extract cross attention and self-attention maps
            
            % \STATE \textbf{// Compute adversarial guidance}
            % \STATE Decode intermediate result: $x_t \leftarrow \mathcal{D}(\tilde{x}_t)$
            \STATE Compute $\mathcal{L}_\mathrm{k}$ via Eq.~\eqref{eq:eq6}~\eqref{eq:eq7}~\eqref{eq:eq9}
            
            % \STATE \textbf{// Score matching with adversarial guidance}
            \STATE Compute adversarial  guided score via Eq.~\eqref{eq:eq11}
            
            % \STATE \textbf{// Denoise with EHS schedule}
            \STATE Update: $\tilde{x}_{t-1}$ via Eq.~\eqref{eq:eq12}
        \ENDFOR
        \ENDFOR
        \STATE Decode $x_{\mathrm{adv}}$ to pixel space
        
        \STATE \textbf{// Phase 2: INN-Based Reversible Steganography}
        \STATE Generate $x_{\mathrm{rae}}$ $\leftarrow$ Embed $x_{\mathrm{orig}}$ into $x_{\mathrm{adv}}$ via \cref{eq:eq13}
        \RETURN $x_{\mathrm{rae}}$
    \end{algorithmic}
\end{algorithm}

%% file: sec/5-experiment.tex
\section{Experiments}
\label{sec:experiments}

\subsection{Experimental Setup}
\definecolor{oursblue}{HTML}{E6F3FF} 

\begin{table*}[t!]
\centering
\caption{Comparison of different privacy-preserving methods and multi-modal adversarial attacks in terms of protection performance across four vision-language models. All metrics are lower-is-better and normalized to 0\,--\,100.}
\setlength{\tabcolsep}{4pt}
\renewcommand{\arraystretch}{1.08}

\begin{minipage}[t]{0.490\textwidth}
    \centering
    \resizebox{\linewidth}{!}{
    \begin{tabular}{lcccccc}
        \toprule[1.2pt]
        \multicolumn{7}{c}{\textbf{(a) BLIP}} \\
        \midrule
        \textbf{Method} & \textbf{BLEU-1} & \textbf{BLEU-4} & \textbf{CIDEr} & \textbf{METEOR} & \textbf{ROUGEL} & \textbf{SPICE} \\
        \midrule
        DP-RAE~\cite{zhu2024dp}              & 81.12 & 67.32 & 64.90 & 77.67 & 79.43 & 73.18 \\
        Anyattack~\cite{zhang2024anyattack} & 70.36 & 49.69 & 48.23 & 65.79 & 69.63 & 53.53 \\
        AttackVLM~\cite{zhao2023evaluating} & 75.43 & 60.13 & 56.55 & 70.89 & 72.94 & 60.38 \\
        AdvDiffVLM~\cite{guo2024efficient}  & 69.72 & 47.37 & 44.88 & 64.44 & 68.23 & 48.36 \\
        \rowcolor{advgray}
        \textbf{CloakDiff-ADV}              & 63.78 & 43.19 & 41.69 & 58.95 & 62.49 & 43.56 \\
        \rowcolor{raeblue}
        \textbf{CloakDiff-RAE}              & 64.25 & 44.10 & 41.91 & 59.17 & 63.23 & 45.18 \\
        \bottomrule[1.2pt]
    \end{tabular}}
\end{minipage}
\hfill
\hspace{0.01\textwidth}
\hfill
\begin{minipage}[t]{0.490\textwidth}
    \centering
    \resizebox{\linewidth}{!}{
    \begin{tabular}{lcccccc}
        \toprule[1.2pt]
        \multicolumn{7}{c}{\textbf{(b) Unidiffuser}} \\
        \midrule
        \textbf{Method} & \textbf{BLEU-1} & \textbf{BLEU-4} & \textbf{CIDEr} & \textbf{METEOR} & \textbf{ROUGEL} & \textbf{SPICE} \\
        \midrule
        DP-RAE~\cite{zhu2024dp}              & 70.56 & 52.76 & 50.05 & 69.33 & 69.45 & 51.29 \\
        Anyattack~\cite{zhang2024anyattack} & 52.24 & 30.51 & 29.01 & 48.52 & 50.02 & 33.07 \\
        AttackVLM~\cite{zhao2023evaluating} & 46.00 & 29.41 & 29.20 & 48.40 & 44.75 & 34.15 \\
        AdvDiffVLM~\cite{guo2024efficient}  & 55.33 & 32.97 & 29.73 & 51.95 & 52.81 & 42.89 \\
        \rowcolor{advgray}
        \textbf{CloakDiff-ADV}              & 35.22 & 15.34 & 12.38 & 30.19 & 32.71 & 14.22 \\
        \rowcolor{raeblue}
        \textbf{CloakDiff-RAE}              & 37.10 & 16.29 & 13.57 & 31.85 & 34.88 & 14.86 \\
        \bottomrule[1.2pt]
    \end{tabular}}
\end{minipage}

\vspace{0.9em}

\begin{minipage}[t]{0.490\textwidth}
    \centering
    \resizebox{\linewidth}{!}{
    \begin{tabular}{lcccccc}
        \toprule[1.2pt]
        \multicolumn{7}{c}{\textbf{(c) LLaVA}} \\
        \midrule
        \textbf{Method} & \textbf{BLEU-1} & \textbf{BLEU-4} & \textbf{CIDEr} & \textbf{METEOR} & \textbf{ROUGEL} & \textbf{SPICE} \\
        \midrule
        DP-RAE~\cite{zhu2024dp}              & 72.48 & 57.39 & 48.92 & 68.15 & 68.88 & 65.84 \\
        Anyattack~\cite{zhang2024anyattack} & 60.73 & 40.02 & 29.34 & 53.13 & 54.00 & 52.89 \\
        AttackVLM~\cite{zhao2023evaluating} & 66.17 & 48.46 & 37.96 & 60.04 & 60.87 & 56.55 \\
        AdvDiffVLM~\cite{guo2024efficient}  & 59.46 & 36.90 & 25.23 & 49.60 & 50.10 & 38.42 \\
        \rowcolor{advgray}
        \textbf{CloakDiff-ADV}              & 53.04 & 30.55 & 21.21 & 44.53 & 45.80 & 32.06 \\
        \rowcolor{raeblue}
        \textbf{CloakDiff-RAE}              & 53.09 & 31.05 & 21.51 & 44.48 & 45.61 & 32.29 \\
        \bottomrule[1.2pt]
    \end{tabular}}
\end{minipage}
\hfill
\hspace{0.01\textwidth}
\hfill
\begin{minipage}[t]{0.490\textwidth}
    \centering
    \resizebox{\linewidth}{!}{
    \begin{tabular}{lcccccc}
        \toprule[1.2pt]
        \multicolumn{7}{c}{\textbf{(d) Qwen}} \\
        \midrule
        \textbf{Method} & \textbf{BLEU-1} & \textbf{BLEU-4} & \textbf{CIDEr} & \textbf{METEOR} & \textbf{ROUGEL} & \textbf{SPICE} \\
        \midrule
        DP-RAE~\cite{zhu2024dp}              & 75.18 & 64.02 & 65.37 & 77.41 & 79.46 & 73.11 \\
        Anyattack~\cite{zhang2024anyattack} & 60.46 & 49.21 & 48.73 & 65.12 & 69.08 & 53.44 \\
        AttackVLM~\cite{zhao2023evaluating} & 75.44 & 60.03 & 56.89 & 70.83 & 72.65 & 60.71 \\
        AdvDiffVLM~\cite{guo2024efficient}  & 69.32 & 47.66 & 45.02 & 64.18 & 68.37 & 48.91 \\
        \rowcolor{advgray}
        \textbf{CloakDiff-ADV}              & 49.15 & 26.64 & 20.28 & 47.01 & 45.99 & 41.03 \\
        \rowcolor{raeblue}
        \textbf{CloakDiff-RAE}              & 51.03 & 27.09 & 21.44 & 47.84 & 46.48 & 42.98 \\
        \bottomrule[1.2pt]
    \end{tabular}}
\end{minipage}
\label{tab:comparison_protection_2x2}
\vspace{-0.3cm}
\end{table*}

\begin{figure*}[!t]
    \centering
    \includegraphics[scale=0.53]{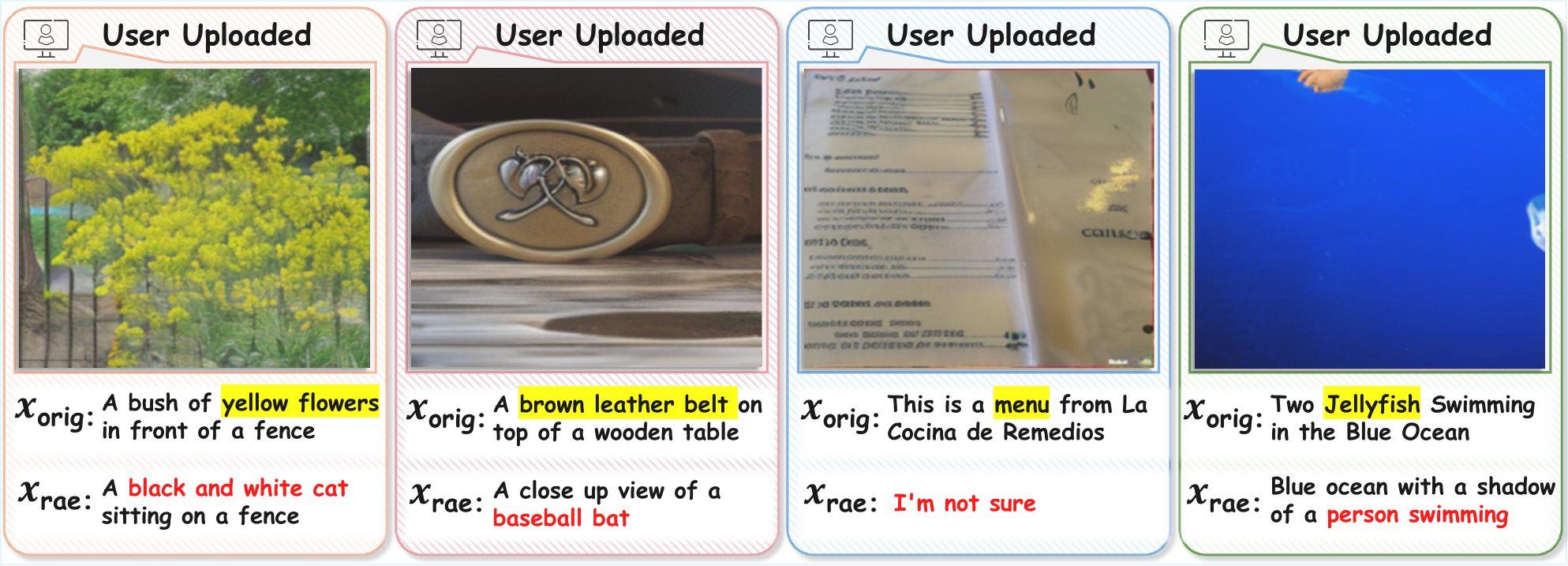}
    % \vspace{-15pt}
    \caption{From left to right: BLIP, UniDiffuser, Flamingo, and LLaVA. Shown are reversible adversarial examples generated by CloakDiff, where "Orig:" and "RAE:" denote outputs from the original and reversible adversarial images, respectively.
    }
    \label{fig:xmodel}
    % \vspace{-0.4cm}
\end{figure*}

%%% 消融实验
\begin{figure*}[!t]   % * 表示忽略单行
\setlength{\abovecaptionskip}{4pt}
  \centering
    \subcaptionbox{Module}{\vspace{2pt}\includegraphics[width=0.24\textwidth]{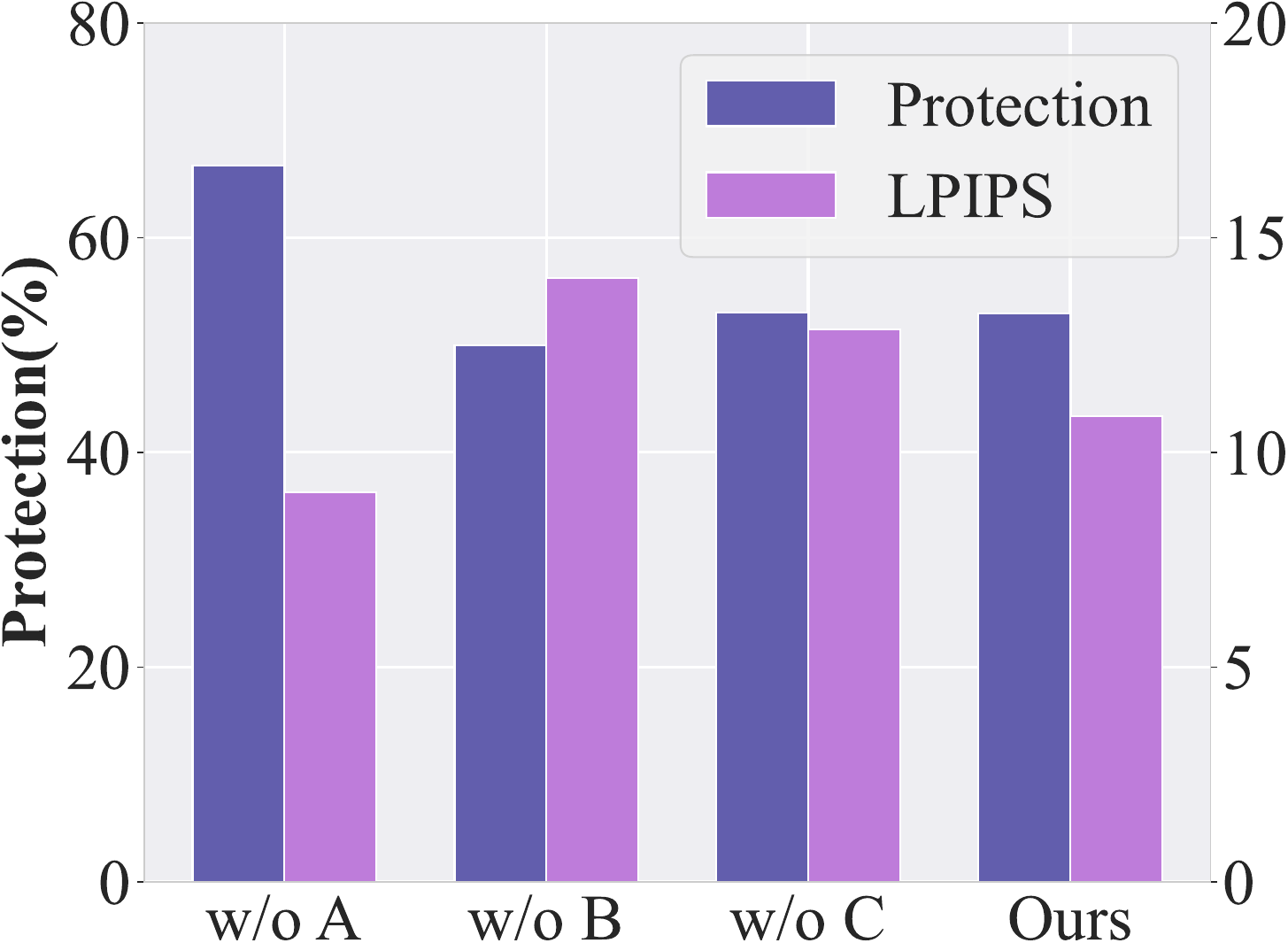}}
    \subcaptionbox{$\lambda_\mathrm{{pixel}}$}{\includegraphics[width=0.24\textwidth]{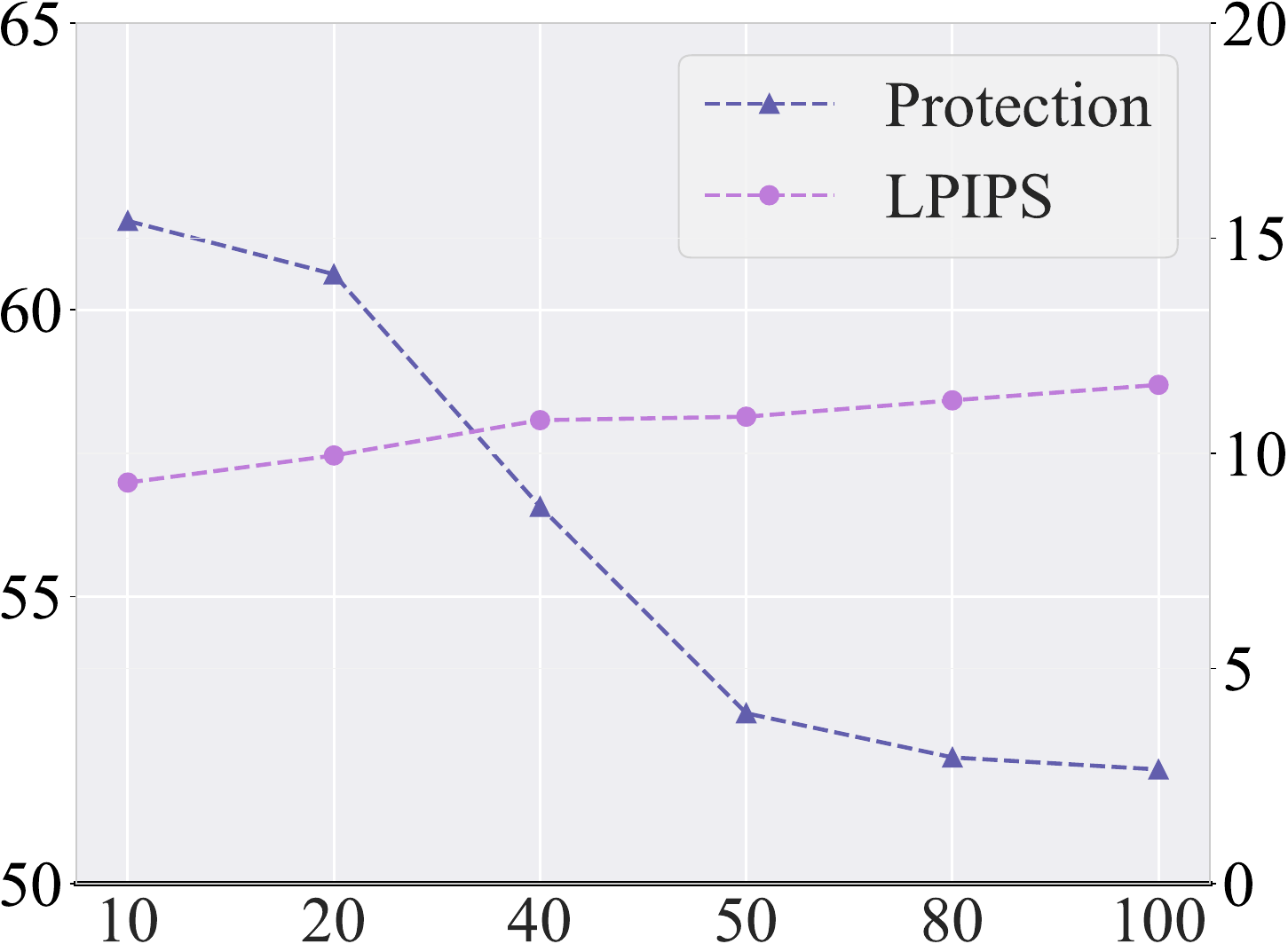}}
    \subcaptionbox{$\lambda_\mathrm{{latent}}$}{\includegraphics[width=0.24\textwidth]{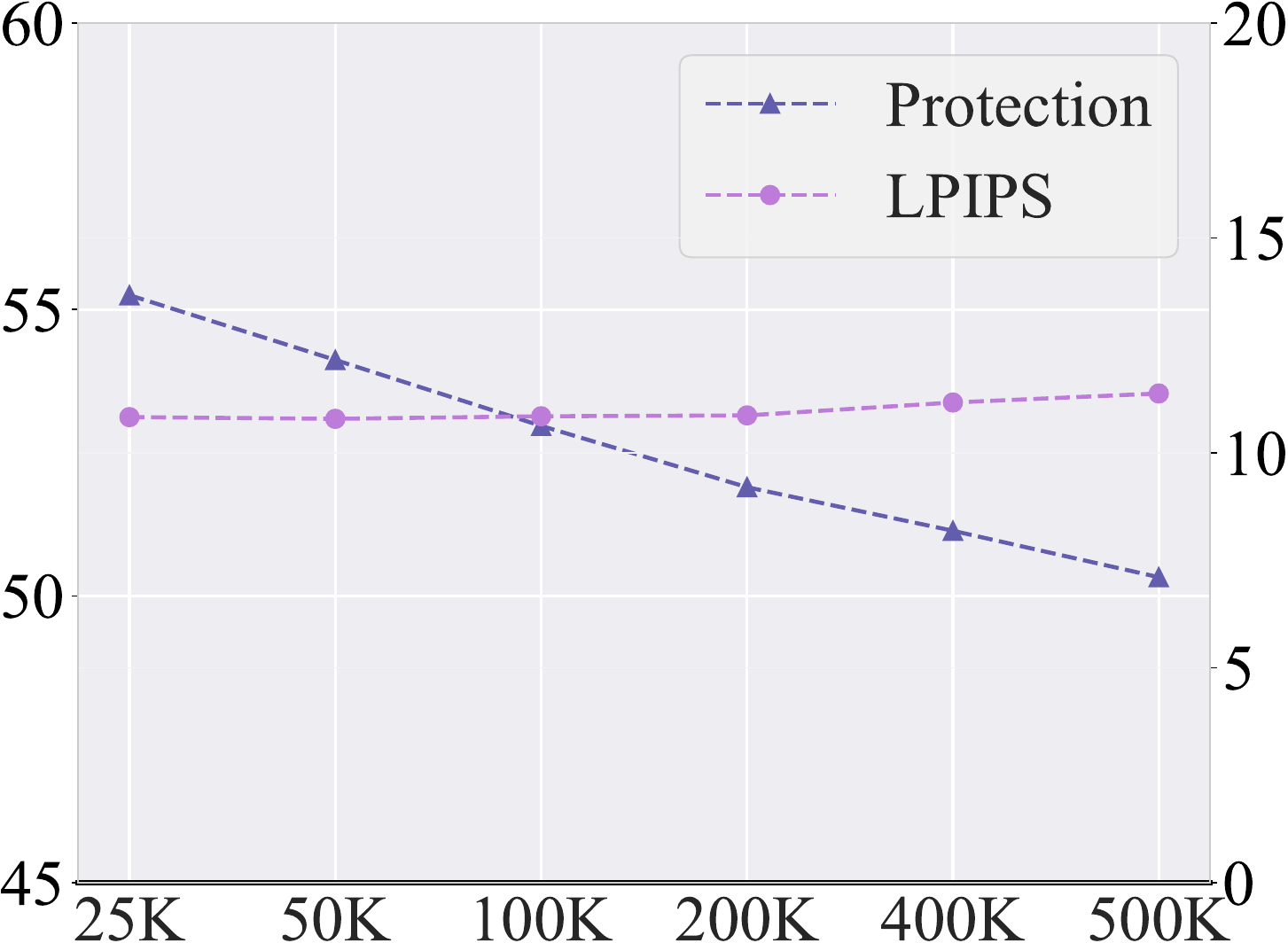}}
    \subcaptionbox{$\lambda_\mathrm{{retain}}$}{\includegraphics[width=0.242\textwidth]{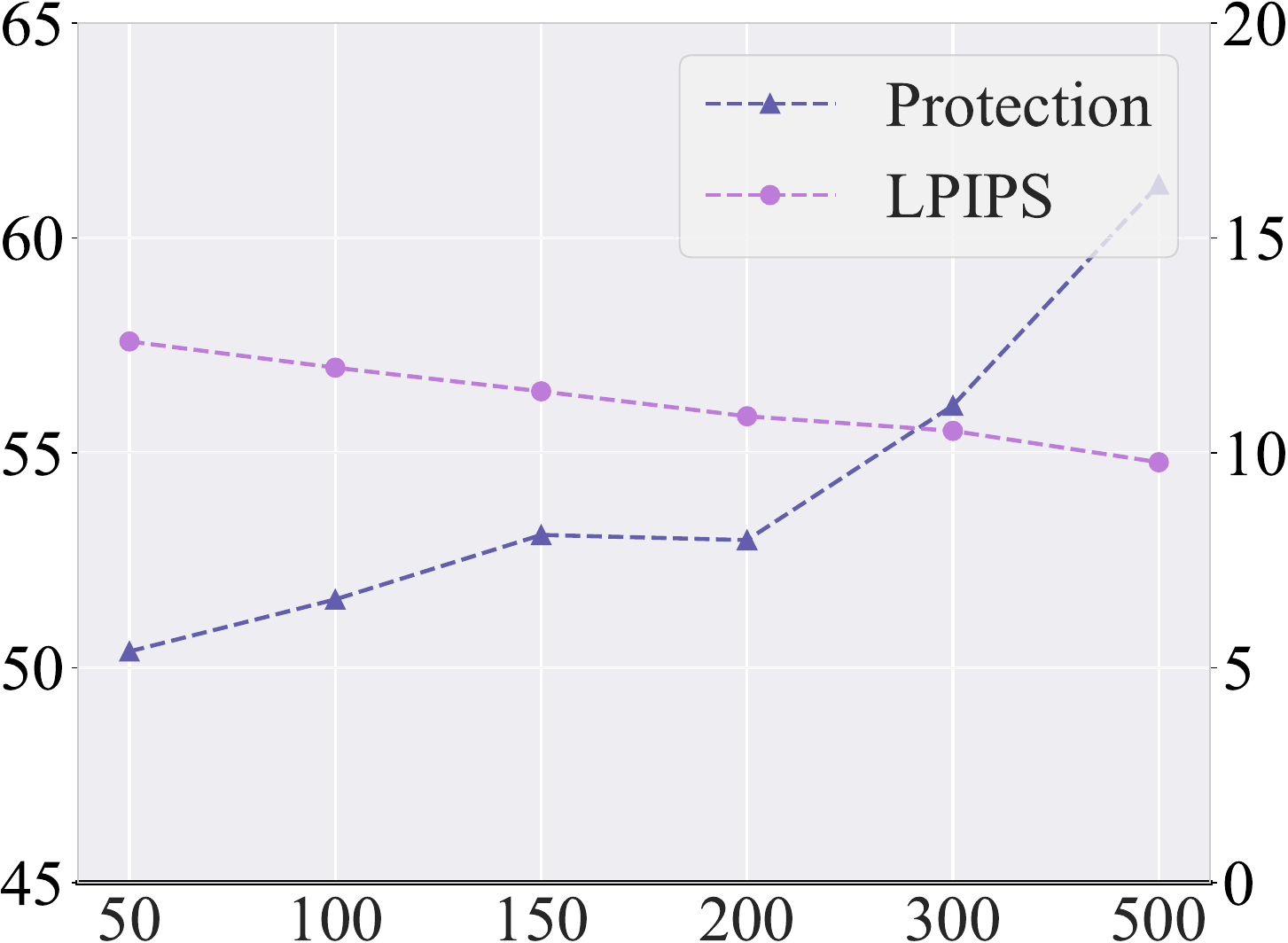}}

    \subcaptionbox{Optimization Step $\alpha_1$}{\includegraphics[width=0.24\textwidth]{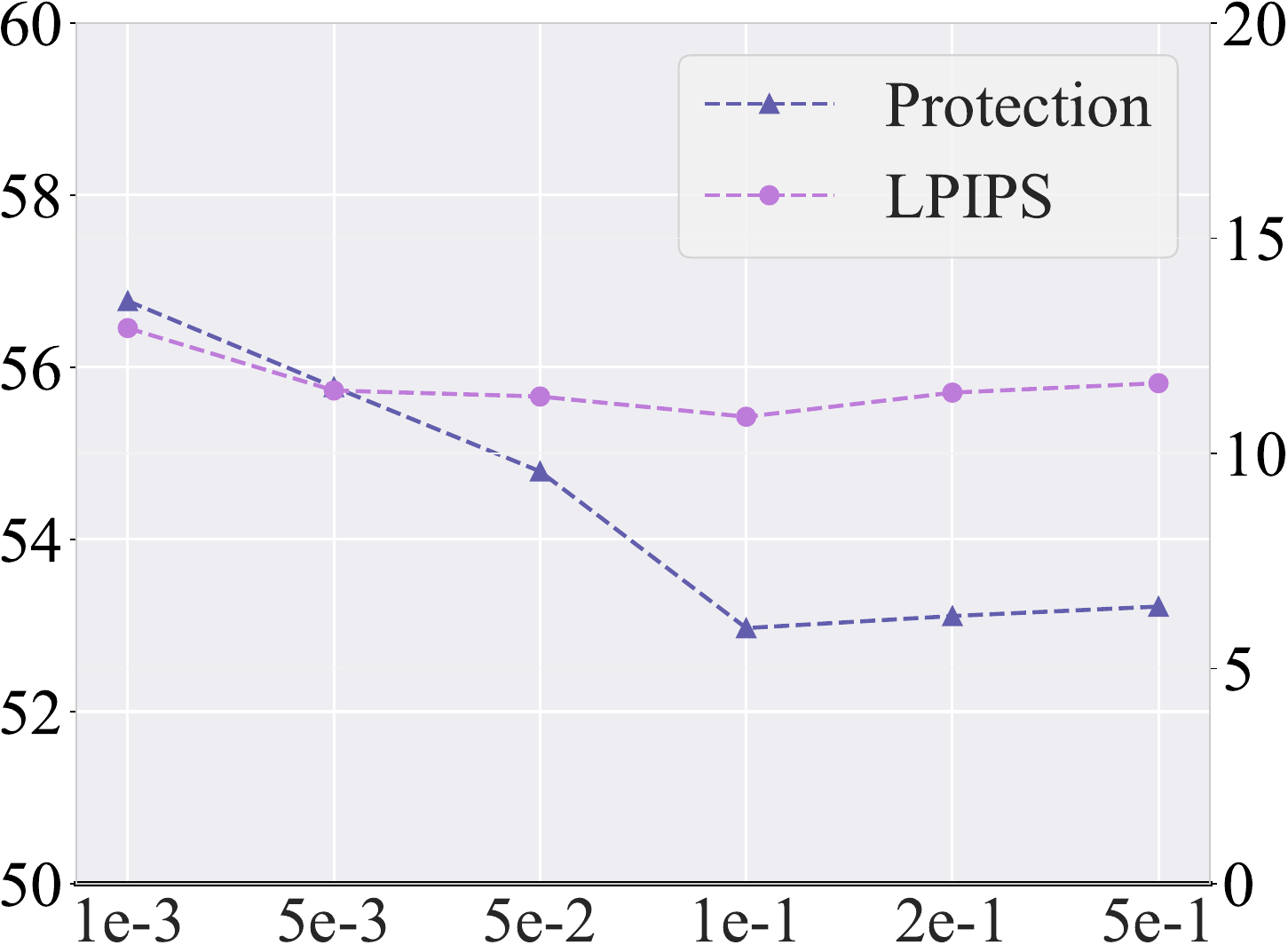}}
    \subcaptionbox{Adversarial Scale $\alpha_2$}{\includegraphics[width=0.24\textwidth]{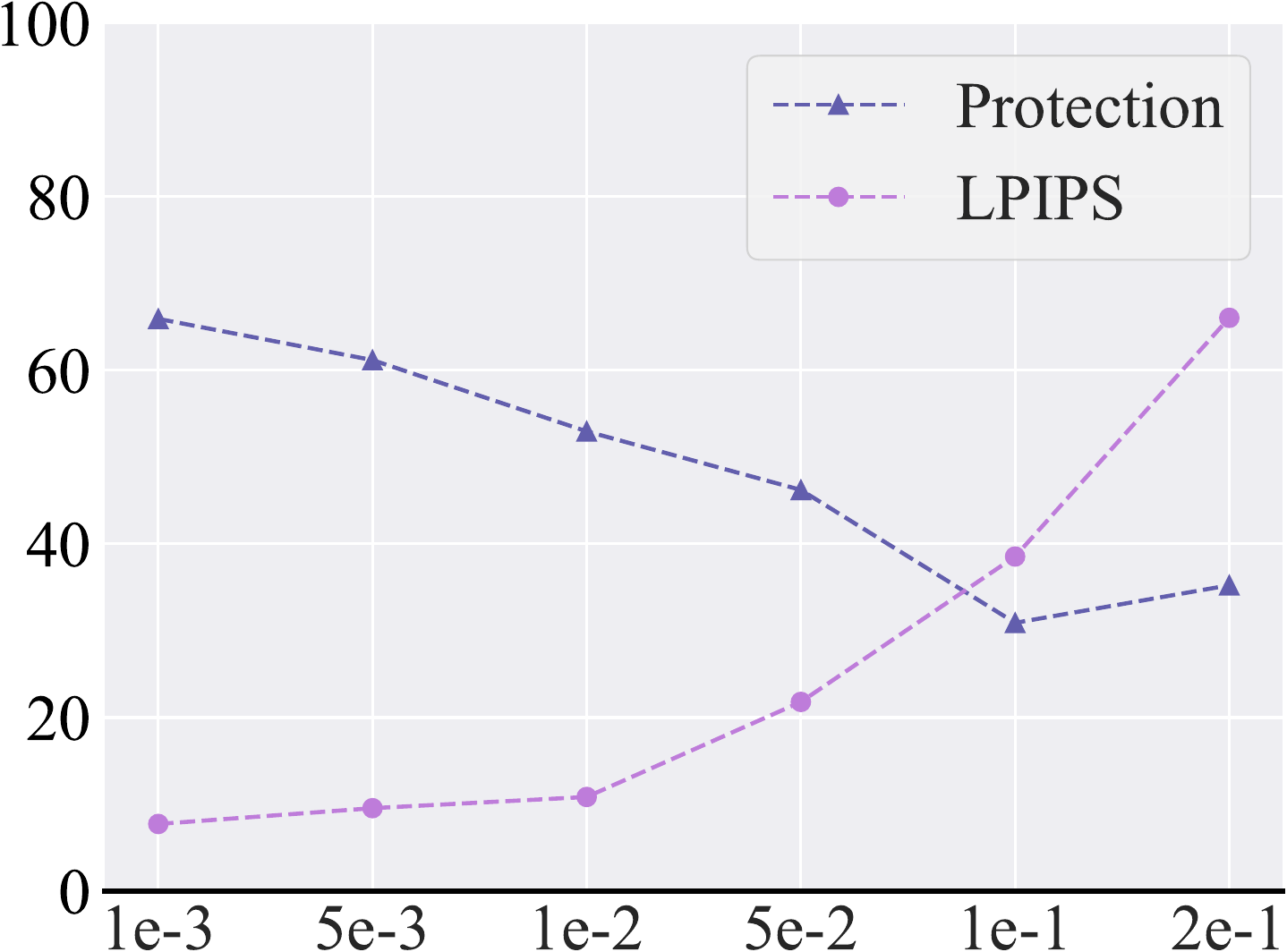}}
    \subcaptionbox{EHS Parameter $\rho$}{\includegraphics[width=0.24\textwidth]{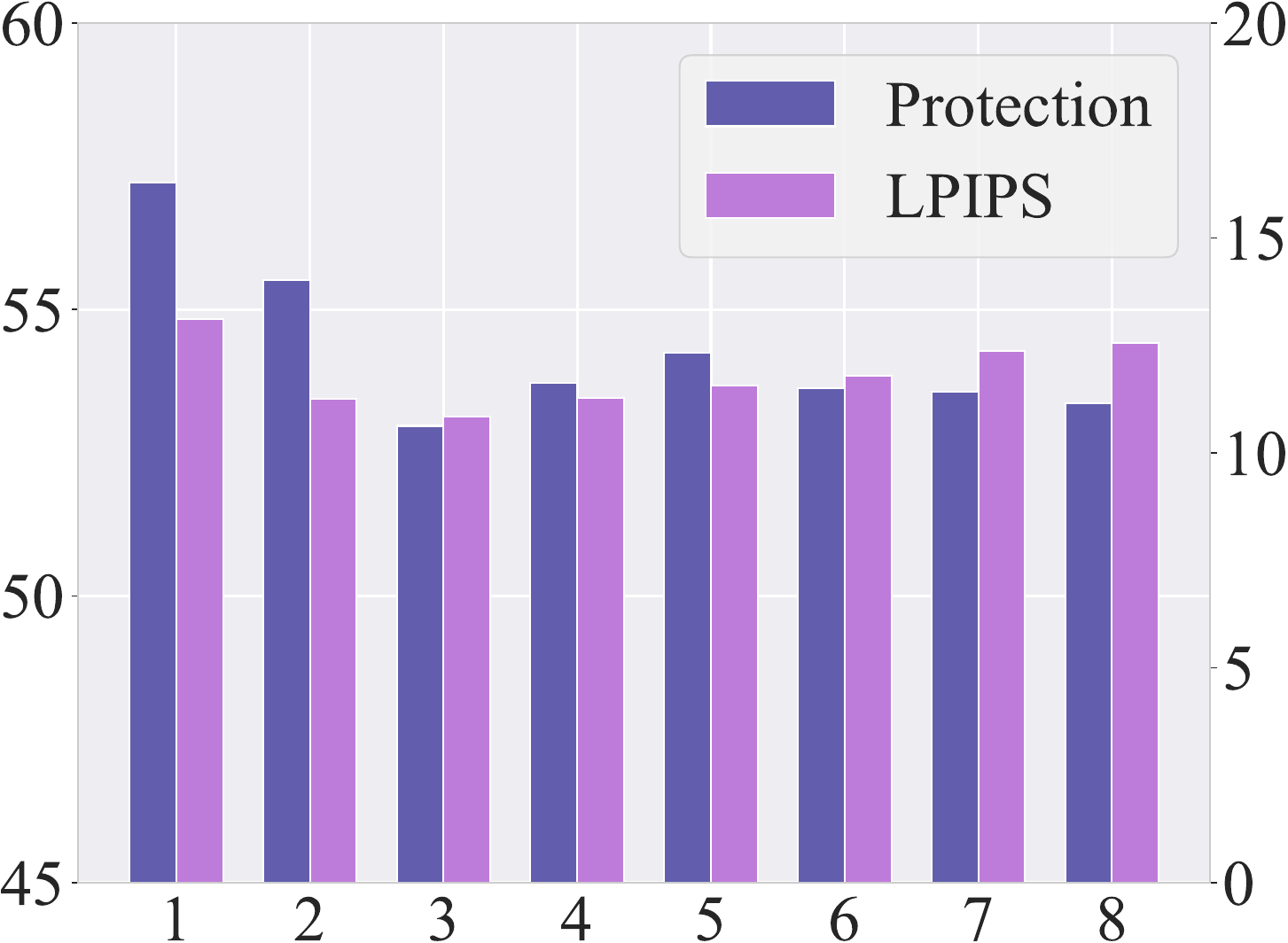}}
    \subcaptionbox{Iteration $K$}{\includegraphics[width=0.24\textwidth]{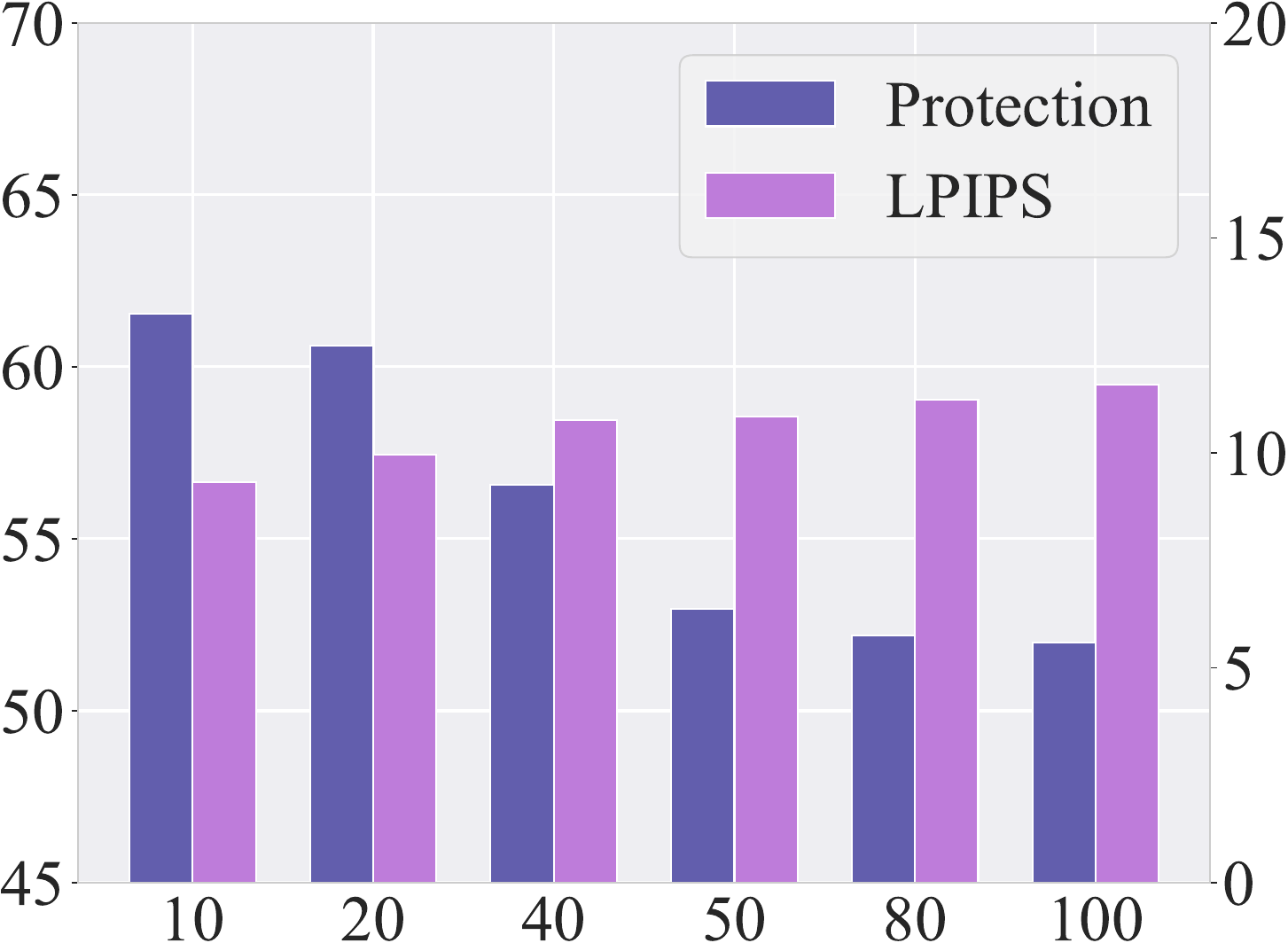}}

    \caption{Ablation study of CloakDiff on BLIP. ‘Protection’ is computed as the average of the six corresponding metrics. The left y-axis represents the value of ‘Protection’, while the right y-axis represents the value of ‘LPIPS’.}
    \label{fig:ablation}
    \vspace{-0.2cm}
\end{figure*}

\noindent\textbf{Models and Datasets.}
We evaluate CloakDiff on two datasets, the ImageNet~\cite{deng2009imagenet} compatible NIPS 2017 adversarial competition dataset~\cite{papernot2016technical} and MS-COCO~\cite{lin2014microsoft}, following~\cite{guo2024efficient} and~\cite{zhang2024anyattack}.
Target images $x_{\mathrm{tgt}}$ are generated using MS-COCO captions as reference texts and synthesized with Stable Diffusion~\cite{rombach2022high}. We assess our method on nine  VLMs covering representative architectures: BLIP~\cite{li2022blip}, BLIP-2~\cite{li2023blip}, InstructBLIP~\cite{dai2023instructblip}, Flamingo~\cite{alayrac2022flamingo}, UniDiffuser~\cite{bao2023one}, MiniGPT-4~\cite{zhu2023minigpt}, LLaVA~\cite{liu2023visual}, Qwen2.5-VL (7B Instruct)~\cite{bai2025qwen2}, and InternVL (3-5-8B)~\cite{wang2025internvl3}. For each $x_{\mathrm{rae}}$, models are evaluated on 10 randomly sampled text prompts $t$ from a candidate pool. Details of the model and datasets can be found in the appendix.

\noindent\textbf{Evaluation Metrics.}
To assess protection performance, we measure the similarity between captions generated by the VLMs for $x_{\mathrm{rae}}$ and $x_{\mathrm{orig}}$, using standard image captioning metrics, 
BLEU$\,\downarrow$, CIDEr$\,\downarrow$, METEOR$\,\downarrow$, ROUGE-L$\,\downarrow$, and SPICE$\,\downarrow$, following~\cite{zhang2024anyattack}. For visual fidelity, we adopt LPIPS~\cite{zhang2018unreasonable}$\,\downarrow$ and FID ~\cite{heusel2017gans}$\,\downarrow$ to assess $x_{\mathrm{rae}}$, following ~\cite{guo2024efficient}. For recovery quality of $x_{\mathrm{orig}} ^ \mathrm{{rev}}$, we use SSIM$\,\uparrow$, PSNR$\,\uparrow$, and RMSE$\,\downarrow$. In all evaluations, $\,\uparrow$ indicates that higher values are better, while $\,\downarrow$ indicates that lower values are better.

\noindent\textbf{Parameter Setting.}
We use Stable Diffusion~\cite{rombach2022high} as the diffusion model and ViT-B/32 as CLIP encoder $f_{\phi}$. The total sampling steps $N$ are set to 20, and the EHS parameter $\rho$ is 3. We set the loss weights to $\lambda_\mathrm{{pixel}}=50$, $\lambda_\mathrm{{latent}}=200$, and $\lambda_\mathrm{{retain}}=100K$ by default.
The optimization step size $\alpha_1$ is set to $1e{-1}$, the adversarial scale $\alpha_2$ is set to $1e{-2}$. And the iteration $K$ is set to 50 by default. 
% A detailed ablation of these parameters is provided in Section \ref{sec:comparison}.

\subsection{Overall Performance}
To evaluate the capability of CloakDiff, we rescale all six metrics to a uniform range of 0\,–\,100. All VLMs generate deterministic text outputs to guarantee reproducibility. As shown in \cref{tab:main}, CloakDiff consistently achieves strong performance across both datasets and all models, leading to a significant reduction in text similarity. The method remains effective even against robust and SOTA models such as Qwen and InternVL, demonstrating the cross-model and cross-prompt protection capability of CloakDiff. For fidelity evaluation of $x_{\mathrm{rae}}$, we adopt two widely used metrics: LPIPS for perceptual similarity and FID for distributional similarity between $x_{\mathrm{orig}}$ and $x_{\mathrm{rae}}$.
As shown in ~\cref{tab:fidelity}, the results confirm that CloakDiff generates $x_{\mathrm{rae}}$ with high visual fidelity. 
During recovery, we evaluate the reconstruction quality of $x_{\mathrm{orig}}^{\mathrm{rev}}$ against $x_{\mathrm{orig}}$. As shown in \cref{tab:fidelity}, the restored images maintain high fidelity, avoiding any irreversible degradation. To further assess the privacy protection capability of CloakDiff, we evaluate it on the Visual Inference Privacy dataset~\cite{vip}. CloakDiff also extends naturally to the setting of explicit attribute inference. As shown in~\cref{tab:bias_attribute_results}, CloakDiff reduces the average performance of VLMs by 45\%, demonstrating strong efficacy in mitigating inference on sensitive attributes. As shown in~\cref{fig:xmodel}, samples protected by CloakDiff exhibit high visual fidelity while substantially disrupting the outputs of VLMs, demonstrating its strong protective efficacy across different model architectures.

\subsection{Comparison Study}
\label{sec:comparison}
To evaluate the effectiveness of CloakDiff as a privacy-preserving framework, we compare it with several representative baselines. We include multi-modal adversarial attack methods such as Anyattack ~\cite{zhang2024anyattack}, AttackVLM ~\cite{zhao2023evaluating}, and reversible adversarial examples (RAE) in multi-modal settings. Since these approaches only generate adversarial examples, we apply the same INN-based reversible steganography method as CloakDiff to obtain the final $x_{\mathrm{rae}}$. And we further compare against the SOTA DP-RAE ~\cite{zhu2024dp}, which adopts a traditional reversible data hiding method. For evaluation, we use ImageNet as the clean image dataset and select three VLMs: BLIP, UniDiffuser, and LLaVA. As shown in ~\cref{tab:comparison_protection_2x2}, CloakDiff-ADV achieves the strongest protection performance. When the original image $x_{\mathrm{orig}}$ is embedded into the adversarial image $x_{\mathrm{adv}}$, CloakDiff-RAE slightly decreases compared to the direct use of $x_{\mathrm{adv}}$. For visual quality, we measure perceptual fidelity between $x_{\mathrm{rae}}$ and $x_{\mathrm{orig}}$, as shown in~\cref{tab:fidelity}. Anyattack and AttackVLM inject high-frequency noise, yielding substantially higher LPIPS and FID. AdvDiffVLM targets imperceptibility but its heavy sampling disruption still produces noticeable degradation. CloakDiff significantly outperforms baseline methods across multiple metrics of visual quality and recovery quality, highlighting its superiority in real world applications.

\subsection{Ablation Study}
\noindent \textbf{Effect of Each Module}:
We consider three modules: (A) Adversarial-Guided Editing, (B) Self-Attention-Based Retention, and (C) EDM-Heuristic Sampling, as shown in \cref{fig:ablation} (a). To assess the effect of module C, we compare it with the default DDIM~\cite{song2020denoising} scheduler.

\begin{table}[t!]
    \caption{Evaluation of CloakDiff on the VIP dataset for five sensitive attributes, including Sex (SEX), Income (INC), Education (EDU), Occupation (OCC), and Marital Status (MAR). Setting A: Users upload $x_{\mathrm{orig}}$, Setting B: Users upload $x_{\mathrm{rae}}$.}
    \vspace{-0.2cm}
    \centering
    % \scriptsize
    \renewcommand{\arraystretch}{1.0}
    \setlength{\tabcolsep}{8pt}
    \resizebox{\columnwidth}{!}{
        \begin{tabular}{ccccccc}
            \toprule[1.5pt]
            \textbf{Model} & \textbf{Setting} & \textbf{SEX} & \textbf{INC} & \textbf{EDU} & \textbf{OCC} & \textbf{MAR} \\
            \midrule
            \multirow{2}{*}{MiniGPT-4} 
                & Setting A   & 0.28 & 0.18 & 0.21 & 0.33 & 0.18 \\
                & Setting B & 0.14 & 0.06 & 0.13 & 0.19 & 0.19 \\
            \midrule
            \multirow{2}{*}{Qwen2.5} 
                &  Setting A  & 0.42 & 0.34 & 0.52 & 0.39 & 0.26 \\
                & Setting B & 0.28 & 0.24 & 0.36 & 0.18 & 0.16 \\
            \midrule
            \multirow{2}{*}{LLaVA} 
                &  Setting A  & 0.32 & 0.36 & 0.43 & 0.31 & 0.17 \\
                & Setting B & 0.14 & 0.15 & 0.38 & 0.07 & 0.12 \\
            \bottomrule[1.5pt]
        \end{tabular}%
    } 
     % \vspace{-0.3cm}
    \label{tab:bias_attribute_results}
    \vspace{-0.7cm}
\end{table}

\noindent \textbf{Effect of the Loss Weights $\lambda$}:
As shown in \cref{fig:ablation} (b)–(d), increasing $\lambda_{\mathrm{pixel}}$ or $\lambda_{\mathrm{latent}}$ improves protection performance but reduces image quality. In contrast, a smaller $\lambda_{\mathrm{retain}}$ can enhance perceptual quality, whereas an overly large value diminishes the adversarial semantics.

\noindent \textbf{Effect of the Unconditional Text Embedding Step Size $\alpha_1$}:
As discussed in \cref{sec:methodology}, improving the reconstruction accuracy of the original image enhances its protection. ~\cref{fig:ablation} (e) shows that a small $\alpha_1$ fails to converge within the given number of steps, whereas a large $\alpha_1$ shifts the initial semantics of the unconditional text.

\noindent \textbf{Effect of the Adversarial Scale $\alpha_2$}:
From a score-matching perspective, $\alpha_2$ controls the relative weighting between adversarial and original scores, thereby determining the strength of adversarial guidance. It effectively acts as the step size for latent perturbation updates. As $\alpha_2$ increases, protection improves but visual fidelity degrades, as shown in \cref{fig:ablation} (f).

\noindent \textbf{Effect of the EHS Parameter $\rho$}:
As described in Theorem 1, under EHS scheduling, the sampling coefficient $\alpha_t$ decreases monotonically with $\rho$, allowing control over the sampling density across different SNR levels through a single parameter. As shown in \cref{fig:ablation} (g), small $\rho$ values lead to excessive early sampling jumps and information loss. Increasing $\rho$ initially improves both protection and perceptual quality, but beyond a certain point, the effects saturate.

\noindent \textbf{Effect of the number of iteration $K$}:
Iteration $K$ controls how often the latent variable is updated. As shown in~\cref{fig:ablation} (h), both protection and image degradation increase monotonically with $K$.

\subsection{Evaluation of EHS}
EHS is not merely a form of hyperparameter search. Compared with existing samplers, including DDIM, EDM \cite{karras2022elucidating}, DPM \cite{lu2022dpm}, and DPM++ \cite{lu2025dpm}, our method consistently achieves superior performance, as shown in \cref{tab:sampler_comparison}. In addition, by adjusting the number of sampling steps, CloakDiff reduces generation time without noticeable degradation in protection performance, as shown in \cref{tab:tradeoff}.

\subsection{Real World Performance}
\begin{figure}[t]
    \centering
{        \includegraphics[width=0.23\textwidth]{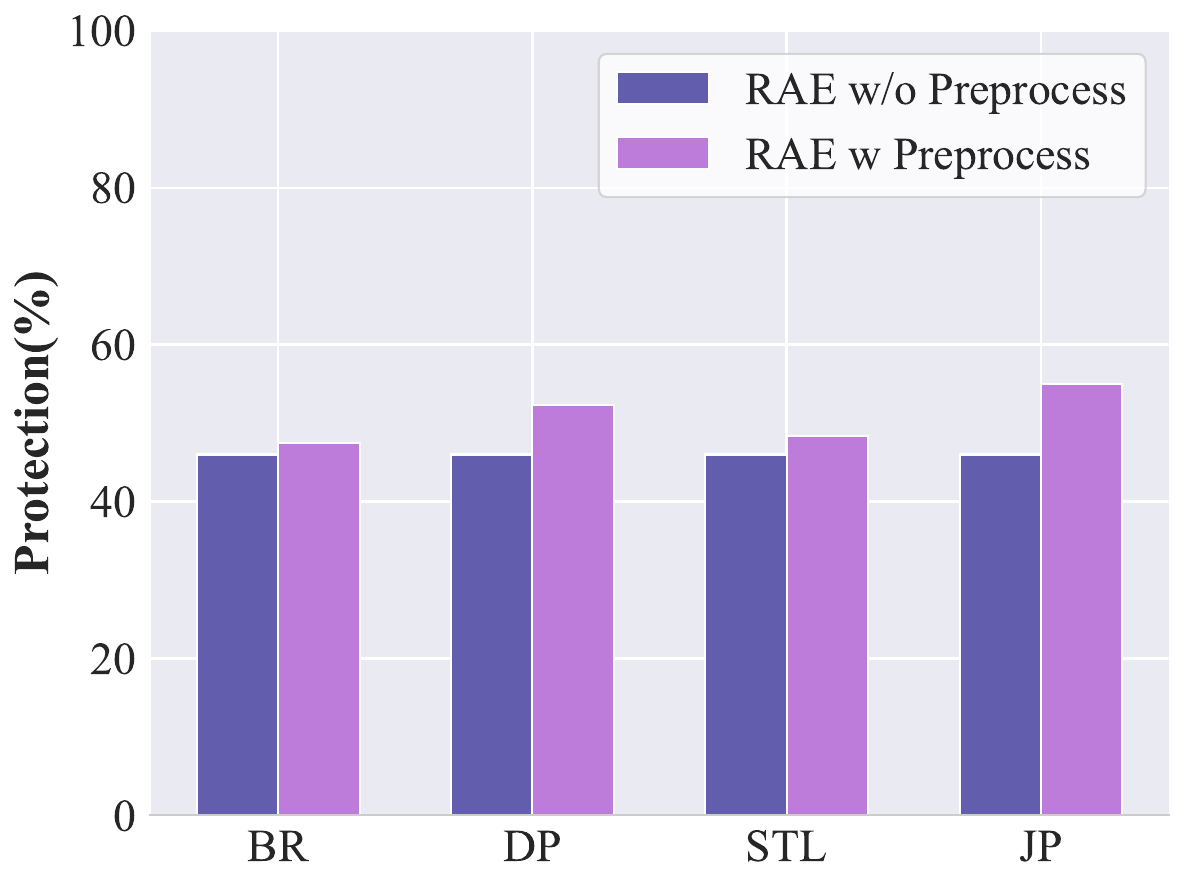}%
    }\hfill
{        \includegraphics[width=0.23\textwidth]{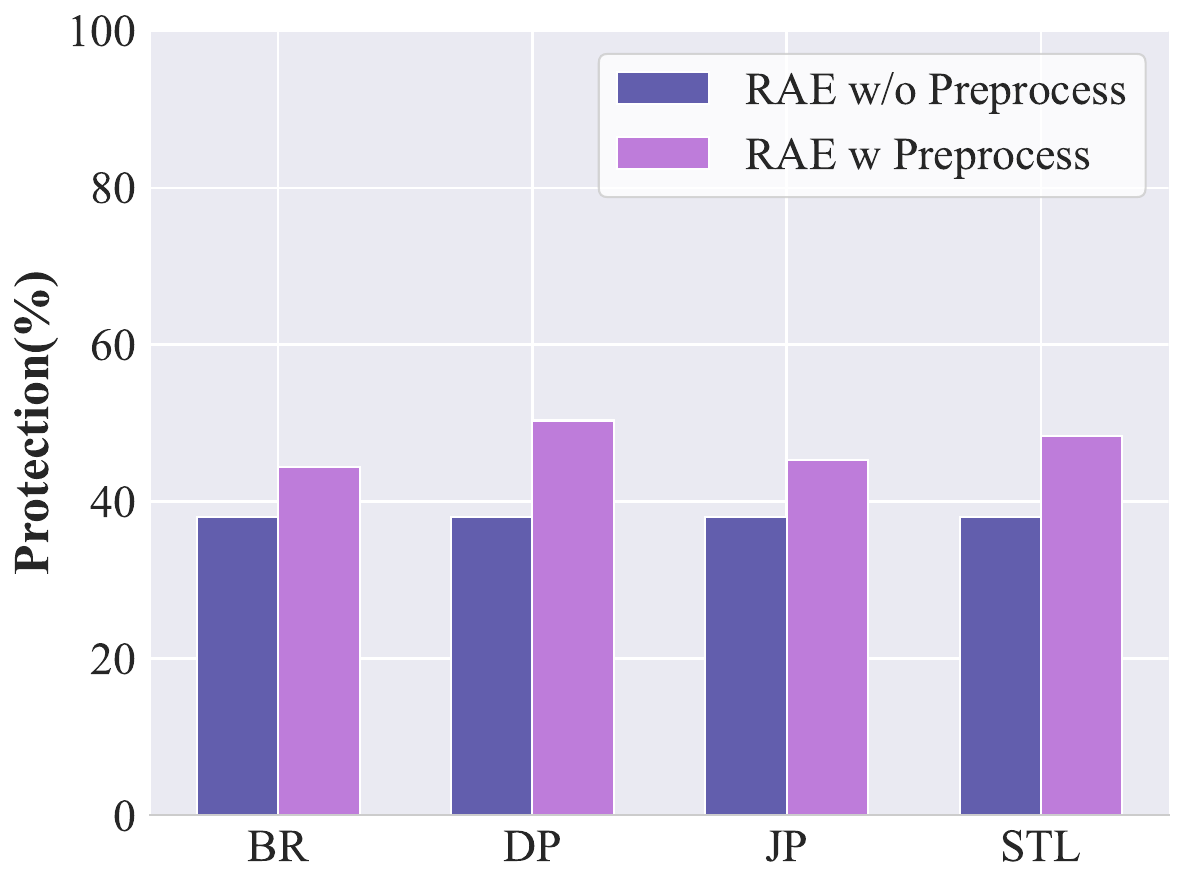}
    }
    \caption{Defense performance of CloakDiff on BLIP (left) and LLaVA (right). ‘w’ denotes with defense, and ‘w/o’ denotes without defense.}
    \label{fig:defense}
    \vspace{-0.3cm}
\end{figure}

In this subsection, we evaluate CloakDiff under real world conditions. We consider a scenario in which users upload images to client applications, and the service provider applies CloakDiff to the uploaded images. In practice, CloakDiff protected images may be subject to compression during network transmission, and attackers may further deploy additional defenses. For transmission induced compression, we consider three widely used formats, PNG~\cite{roelofs2002png}, JPEG~\cite{wallace2002jpeg}, and WebP~\cite{ozturk2021performance}. 
As shown in \cref{tab:compression_recovery}, CloakDiff still enables high quality recovery, although visual quality degrades slightly. We further examine whether common preprocessing defenses can suppress adversarial perturbations and thereby circumvent CloakDiff. We evaluate four representative methods on BLIP and LLaVA. Bit Reduction squeezes features by lowering pixel bit depth. DiffPure maps images into the latent space and removes adversarial noise through diffusion based denoising. STL projects images onto a low dimensional manifold before inference. We also consider network compression as an additional perturbation to the protected images. As shown in \cref{fig:defense}, these defenses have only a marginal effect on the protection offered by CloakDiff.

\begin{table}[t!]
\caption{Protection and Time Trade-off under Different Steps.}
\vspace{-0.3cm}
\centering
\renewcommand{\arraystretch}{0.7}{
\scriptsize
\resizebox{\columnwidth}{!}{%
\begin{tabular}{ccccc}
\toprule[0.8pt]
\textbf{Sampling Steps} & 20 & 15 & 10 & 8 \\
\midrule
Protection Performance & 36.21 & 39.82 & 43.43 & 44.87 \\
Time Cost & 19.08s & 14.26s & 11.07s & 9.32s \\
\bottomrule[0.8pt]
\end{tabular}
}
}
\label{tab:tradeoff}
\vspace{-0.3cm}
\end{table}

\begin{table}[t!]
\caption{Quantitative comparison under different samplers.}
\vspace{-0.2cm}
\centering
\resizebox{\columnwidth}{!}{%
\begin{tabular}{llccccc}
\toprule[1.5pt]
 & & \textbf{EHS (Ours)} & \textbf{DDIM} & \textbf{EDM} & \textbf{DPM} & \textbf{DPM++} \\
\midrule
\multicolumn{2}{c}{LPIPS} & 10.85 & 13.12 & 13.52 & 14.09 & 13.27 \\
\midrule
\multirow[c]{3}{*}{\makecell[c]{Protection\\Performance}} & MiniGPT-4 & 23.71 & 25.88 & 27.35 & 26.52 & 26.05 \\
 & Qwen2.5 & 39.48 & 43.96 & 41.17 & 48.97 & 44.31 \\
 & LLaVA & 38.01 & 44.48 & 39.49 & 43.21 & 43.57 \\
\bottomrule[1.5pt]
\end{tabular}%
}
\label{tab:sampler_comparison}
\vspace{-0.3cm}
\end{table}

\begin{table}[t!]
\caption{Recovery quality of $x_{\mathrm{orig}}^\mathrm{rev}$ under different compression schemes in real world network environments.}
\vspace{-0.3cm}
    \centering
\renewcommand{\arraystretch}{0.8}
    \setlength{\tabcolsep}{8pt}
    \resizebox{\columnwidth}{!}{%
        \begin{tabular}{ccccc}
            \toprule[1.5pt]
            \textbf{Metric} & \textbf{w/o Compression} & \textbf{PNG} & \textbf{JPEG} & \textbf{WebP} \\
            \midrule
            SSIM  & 0.99   & 0.94   & 0.83   & 0.94 \\
            PSNR  & 44.37  & 36.35  & 30.29  & 36.35 \\
            RMSE  & 0.0004 & 0.0639 & 1.0723 & 0.0741 \\
            \bottomrule[1.5pt]
        \end{tabular}%
        }
         % \vspace{-0.3cm}
                \label{tab:compression_recovery}
        \vspace{-0.4cm}
\end{table}

%% file: sec/6-conclusion.tex
\section{Conclusion and Limitation}
\label{sec:Conclusion}
In this paper, we present CloakDiff, the first framework for privacy protection against text-based query attacks in VLM-driven social media. It generates imperceptible yet reversible adversarial examples by combining diffusion-based adversarial editing with an invertible network for lossless recovery. CloakDiff perturbs pixel embeddings and latent cross attention map to induce fine-grained multi-modal semantic distortion. 
Adversarial cues are progressively injected along the sampling trajectory, while self-attention retention loss preserves global structure and visual fidelity.
We further design a tailored adversarial-guided diffusion schedule with a theoretical foundation in score matching.
% While CloakDiff is highly effective for images, its applicability to videos with continuous spatiotemporal semantics remains to be investigated. 
While CloakDiff is effective for images, its extension to videos with continuous spatiotemporal semantics remains unknown.
As short-video sharing becomes increasingly popular on social platforms, extending CloakDiff to this domain represents a promising direction for future research.

% To address this, we propose CloakDiff—the first privacy-preserving framework against text-based query attacks in VLM-driven social media scenarios. CloakDiff generates imperceptible yet reversible adversarial examples via a diffusion-based editing process, and employs an invertible neural network (INN) for reversible steganography.

% During adversarial example generation, CloakDiff perturbs high-dimensional pixel-space features and disrupts cross-attention in latent variables to achieve fine-grained multi-modal semantic distortion. Through adversarial-guided editing, adversarial information is progressively injected into the sampling trajectory. Meanwhile, a self-attention retention loss constrains global structure to enhance perceptual fidelity. We further design a sampling schedule tailored for adversarial-guided diffusion, and derive its theoretical formulation from a score-matching perspective.

% While our study focuses on image-level privacy in VLM scenarios, a potential limitation is that the protection effectiveness of CloakDiff on videos with continuous frame semantics remains unclear. As short-video sharing becomes increasingly popular on social platforms, extending CloakDiff to spatiotemporal domains represents an important direction for future research.